\documentclass[preprint,3p]{elsarticle}
\usepackage{hyperref}
\usepackage{amssymb, graphicx, mathtools, caption, subcaption, algorithm, kotex}
\usepackage[onehalfspacing]{setspace}
\usepackage[overload]{empheq}
\usepackage[noend]{algpseudocode}
\usepackage{textcomp}

\usepackage{xcolor}

\usepackage{amsthm}
\newtheorem*{remark}{Remark}

\usepackage{multirow,tabularx}

\newcommand{\AL}{CEEN}
\newcommand{\ALtxt}{\text{CEEN}}

\biboptions{sort&compress}
\begin{document}

\begin{frontmatter}

\title{CEENs: Causality-enforced evolutional networks for solving time-dependent partial differential equations}
\author[1]{Jeahan Jung\fnref{eq}}
\ead{obok13@postech.ac.kr}
\author[1]{Heechang Kim\fnref{eq}}
\ead{heechangkim@postech.ac.kr}
\author[1]{Hyomin Shin}
\ead{zhainl@postech.ac.kr}
\author[1]{Minseok Choi\corref{cor}}
\ead{mchoi@postech.ac.kr}

\cortext[cor]{Corresponding author}
\fntext[eq]{Authors contributed equally}
\address[1]{Department of Mathematics, Pohang University of Science and Technology (POSTECH), Pohang 37673, Republic of Korea}
\footnotetext{This is the accepted manuscript of an article published in \textit{Computer Methods in Applied Mechanics and Engineering}, Vol. 427, 117036 (2024). The final published version is available at https://doi.org/10.1016/j.cma.2024.117036.}
\begin{abstract}

Despite the growing popularity of physics-informed neural networks (PINNs), 
their applicability in the long-time integration of partial differential
equations (PDEs) remains constrained.
We argue that this problem stems from the lack of consideration of temporal causality in the original PINN
formulation, resulting in a bias towards satisfying governing equations at
later times before learning the initial condition and hence leading to erroneous
solutions. To this end, we propose a novel method that seamlessly integrates
temporal causality into the training process. 
Drawing inspiration from classical numerical methods where the temporal
causality is reflected, we divide the time domain
into nonoverlapping subintervals, assign a unique neural network to
each subinterval, and construct a loss function founded on the integral form of PDEs
within these subintervals. The proposed networks undergo sequential training,
beginning with the initial time step.
Our method demonstrates significant improvement in accuracy for long-time
simulations of various PDE problems where the original PINN method fails
while it requires less computational cost and memory compared to the PINN
method. A parallelization algorithm is provided to further enhance the
computational efficiency, showing a significant speedup for solving
time-dependent PDEs.

\end{abstract}


\end{frontmatter}

%--------------------------------------------------------------
%
% SECTION: INTRODUCTION
%
%--------------------------------------------------------------
\section{Introduction}\label{sec:intro}

The tremendous growth of machine learning technology has led to explosive
advancement in various fields such as natural language processing, data mining,
and computer vision. In particular, physics-informed neural networks (PINNs)
\cite{raissi2019physics,karniadakis2021physics,wang2021understanding,wang2022and,wang2021eigenvector}
have been proposed and become a new trend of solving partial differential
equations (PDEs) using machine learning. The effectiveness of PINNs has been
demonstrated in a variety of PDE problems including the forward and inverse
problems
\cite{raissi2019physics,meng2020composite,jagtap2020conservative,yang2021b,zhang2019quantifying},
and uncertainty quantification
\cite{yang2021b,zhang2019quantifying,yang2020physics,guo2022normalizing,jung2024bayesian,zhong2023pi,karumuri2020simulator,zhang2020learning,cho2022mgdgan,shin2023physics}. The use of neural networks for PDE has also been proposed three decades ago \cite{lagaris1998artificial,dissanayake1994neural}, but use of large-scale neural networks was computationally prohibitive at the time. Advances in computer hardware such as graphic processing units (GPUs) and recent Python libraries such as Tensorflow \cite{tensorflow2015-whitepaper}, Pytorch \cite{NEURIPS2019_9015}, and JAX \cite{jax2018github} provide tools for machine learning-based PDE solver.

Despite the notable achievements of PINNs in solving PDEs, there remains a
significant challenge in integrating time-dependent PDEs given initial
conditions
\cite{mattey2022novel,wight2020solving,wu2022physics,wang2022long,wang2022respecting,krishnapriyan2021characterizing,du2021evolutional}.
Time-dependent PDEs frequently arise in various science and engineering models;
therefore, it is important to understand the underlying pathologies of PINNs in
handling such situations and enhance their efficacy in addressing the
challenges. The literature has highlighted that a major obstacle of PINNs in
solving time-dependent PDEs is the lack of consideration of temporal causality
in the training process. 
This means that PINNs tend to be biased towards satisfying governing equations
at later times before even learning the initial condition, leading to an
inaccurate solution while having low training loss.
To address this pathology, a
number of variants of PINNs have been proposed to incorporate temporal causality
during training. In
\cite{wight2020solving,wang2022respecting,krishnapriyan2021characterizing},
sequential learning strategies of PINN have been proposed to learn the solution
from early time steps. The authors in
\cite{wu2022physics,zheng2022physics,stevens2020finitenet,ren2022phycrnet,hu2020neural,geneva2020modeling,qu2022learning}
employ recurrent neural networks, which are known to effectively handle time
series data, to capture temporal causality. In
\cite{wang2022long,shit2019implicit,li2021physics}, the authors train operators
that predict a solution after a certain period of time from the current state.
The iterative application of these operators, starting from the initial
condition, is used to determine a solution at a specific time. The methods
proposed in \cite{mattey2022novel,wight2020solving,dong2021local,meng2020ppinn}
divide the time domain into short intervals and train a solution in each
interval. The authors in \cite{raissi2019physics,du2021evolutional} assign a
neural network at each time step and train them sequentially. The loss functions
of the assigned neural networks are defined using the Runge-Kutta method, widely
used method of integrating ordinary differential equations (ODEs).

Taking cues from classical numerical methods that naturally respect temporal
causality, we propose a method involving time-domain decomposition and enforcing
sequential learning from the initial time. Our method divides the time domain
into non-overlapping subintervals, assigning a unique neural network to each
subinterval, and constructs a loss function based on the integral form of PDEs
within these subintervals. Sequential training is enforced, beginning with the
initial condition and proceeding through each subsequent subinterval. Although
similar to the discrete PINN method based on the Runge-Kutta method in~\cite{raissi2019physics}, our
approach offers key distinctions and advantages, including the construction of a
loss function using a linear multi-step method that allows for parallelization
by avoiding the need for intermediate step calculations. In particular, we use
the trapezoidal rule for the loss function, known for its second-order accuracy
in terms of the time step. In addition we introduce a parallelization algorithm that
can significantly accelerate the training for the long-time integration of PDEs.
Our main contributions can be summarized as:
\begin{itemize}
    \item{We propose a novel method to address the problem of temporal causality
        for time-dependent PDEs by dividing the time domain into subintervals,
        each equipped with a neural network, and enforcing a sequential training
        that initiates from the very initial time step and proceeds through each
        subsequent subinterval.}
    \item{We provide a parallelization algorithm designed to
        expedite the training process, providing a substantial
        speedup and enhanced computational efficiency.}
    \item{We demonstrate the effectiveness of our model for learning long-time
        integration of PDEs where the original PINN method fails.}
\end{itemize}

The rest of the paper is organized as follows. In Section \ref{sec:pinn}, we
introduce the original formulation of PINN proposed in \cite{raissi2019physics}
and explain why it fails to reflect temporal causality in the training process.
To address this limitation, we propose a new model in Section \ref{sec:tren}
that effectively incorporates temporal causality and describe its training
process. In addition, we elaborate on practical considerations to improve the
efficiency of our proposed method and provide the algorithm. To evaluate the
effectiveness of the proposed method, Section \ref{sec:numerical} presents
state-of-the-art results in a set of challenging time-dependent PDE problems
where the original PINN method fails to solve. In Section \ref{sec:conclusion},
we conclude the paper with a brief summary, some limitations, and future
research directions.

%--------------------------------------------------------------
%
% SECTION: PINNs
%
%--------------------------------------------------------------
\section{Physics-informed neural networks (PINNs)}\label{sec:pinn}

In this section, we briefly introduce an overview of PINNs
\cite{raissi2019physics} for solving PDE problems. We consider the
time-dependent PDEs of the form:
\begin{align}
u_t + \mathcal{N}[u] &= 0, t \in [0,T], x \in \Omega \label{eqn:pde} \\ 
u(0,x) &= g(x), x \in \Omega \label{eqn:ic} \\ 
\mathcal{B}[u] &= 0,t \in [0,T], x \in \partial \Omega \label{eqn:bc}
\end{align}
where $\mathcal{N}$ and $\mathcal{B}$ are differential and boundary operators,
respectively, $\Omega \subset \mathbb{R}^d$ is a spatial domain with a boundary
$\partial \Omega$, and $u$ is the solution of the equation. 
We assume that the initial condition $g$ is a smooth function. Note that in the case
where $g$ is a nonsmooth function, there are methods to effectively approximate
such nonsmooth functions \cite{gk20_adaptiveact, TSENG23_nonsmooth}.
In the PINN formulation, the solution $u(t,x)$ is represented by a neural network model
$U(t,x;\theta)$ with its trainable parameters $\theta$. The loss function is
designed to ensure that the model output satisfies 
the initial, boundary condition, and the governing equation, i.e.,
%\eqref{eqn:pde}, \eqref{eqn:ic}, and \eqref{eqn:bc}, i.e.,
\begin{equation*}
\mathcal{L}^{\text{PINN}}(\theta) = \mathcal{L}^{\text{PINN}}_{ic}(\theta) + \mathcal{L}^{\text{PINN}}_{bc}(\theta) + \mathcal{L}^{\text{PINN}}_{res}(\theta)
\end{equation*}
where
\begin{equation}
\label{eqn:pinn_losses}
\begin{aligned}
\mathcal{L}^{\text{PINN}}_{ic}(\theta) &= \frac{1}{N_{ic}}\sum_{j=1}^{N_{ic}}|U(0,x_{ic}^j;\theta) - g(x_{ic}^j)|^2, \\
\mathcal{L}^{\text{PINN}}_{bc}(\theta) &= \frac{1}{N_{bc}}\sum_{j=1}^{N_{bc}}|\mathcal{B}[U](t_{bc}^j,x_{bc}^j;\theta)|^2, \\
\mathcal{L}^{\text{PINN}}_{res}(\theta) &= \frac{1}{N_{res}}\sum_{j=1}^{N_{res}}|U_t(t_{res}^j,x_{res}^j;\theta) + \mathcal{N}[U](t_{res}^j,x_{res}^j;\theta)|^2.
\end{aligned}
\end{equation}
$\{x_{ic}^j\}_{j=1}^{N_{ic}}$, $\{t_{bc}^j,x_{bc}^j\}_{j=1}^{N_{bc}}$, and
$\{t_{res}^j, x_{res}^j\}_{j=1}^{N_{res}}$ denote sampled points to approximate
$L^2$ norm on $\Omega$, $[0,T] \times \partial \Omega$, and $[0,T] \times
\Omega$, respectively. The ultimate goal of PINNs is to determine $\theta$ that
minimizes the loss function $\mathcal{L}^{\text{PINN}}(\theta)$, which is
typically achieved by stochastic gradient-based optimization algorithms. Note that the
temporal and spatial gradients of the model output included in
\eqref{eqn:pinn_losses} can be efficiently calculated by automatic
differentiation.

However, this formulation of PINN may not incorporate temporal
causality in the training process. As demonstrated in \cite{wang2022respecting},
PINNs trained via stochastic gradient descent are implicitly biased towards
approximating the solution to PDEs at later time steps first, even prior to
learning the initial conditions, hence violating temporal causality. 
Consequently, PINNs often fail to solve time-dependent PDEs, despite having
low training loss.

\begin{figure}[H]
    \centering
    \begin{subfigure}{0.9\linewidth}
        \centering \includegraphics[height=4.5cm]{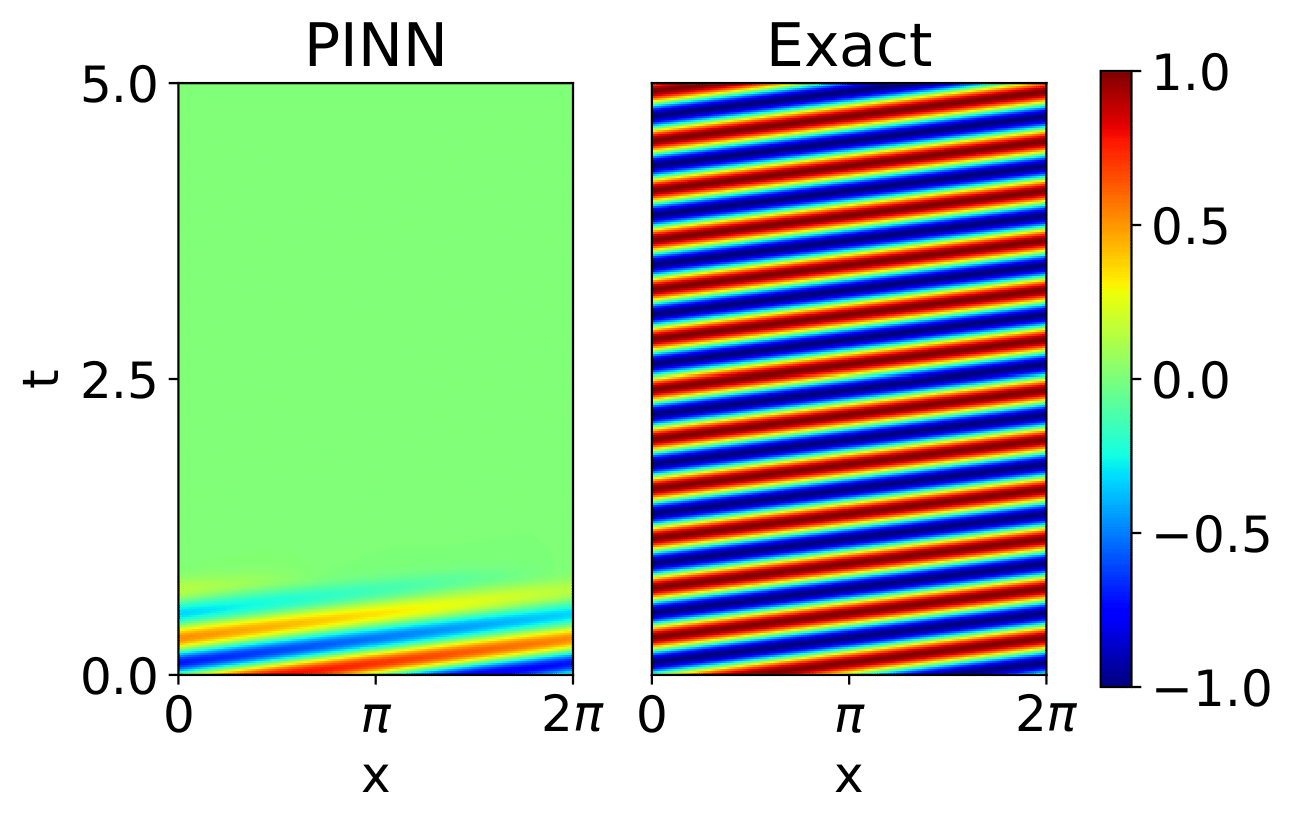}
        \caption{}
    \end{subfigure}
    \begin{subfigure}{0.49\linewidth}
        \centering \includegraphics[height=4.5cm]{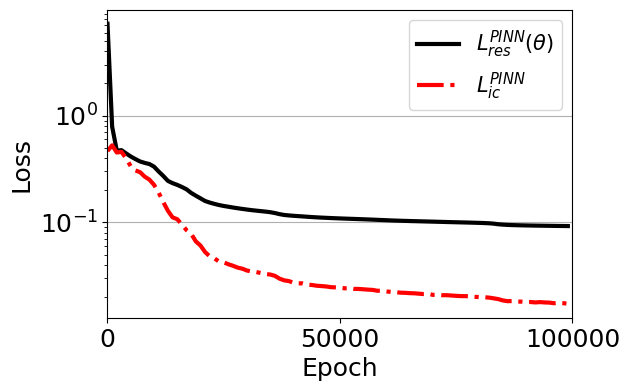}
        \caption{}
    \end{subfigure}
    \begin{subfigure}{0.49\linewidth}
        \centering \includegraphics[height=4.5cm]{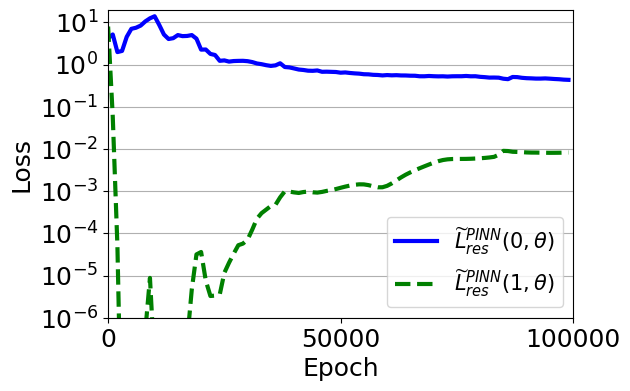}
        \caption{}
    \end{subfigure}
    \caption{PINN performance for the one-dimensional convection equation. (a)
    The prediction of a trained conventional PINN versus reference solution. The
    PINN model provides an inaccurate prediction. (b) Initial
    condition loss of PINN and residual loss of PINN during training. (c) Residual losses of PINN at $t=0$ and 1. 
    The residual loss of PINN at $t=1$ remains low, while that of PINN at
    $t=0$ remains large. This suggests a violation of temporal causality in the
    learning process.}
    \label{fig:convection_pinn}
\end{figure}

To provide a further clarification, we consider the following one-dimensional convection equation.
\begin{equation*}
\begin{aligned}
u_t + 15 u_x &= 0,\;\;\text{for $t \in [0,5], x \in[0,2\pi],$} \\
u(0,x) &= \sin x, \\
u(t,0) &= u(t,2\pi).
\end{aligned}
\end{equation*}
We employ a fully-connected network $U(t,x;\theta)$ with a hyperbolic tangent
activation function to represent the solution $u(t,x)$. The structure of the
network includes 2 hidden layers, and 820 neurons in each hidden layer.
We enforce periodic boundary conditions as hard constraints by
utilizing a suitable Fourier features embedding of the input as introduced
in~\cite{dong2021method}. It ensures that the periodic boundary
condition is directly integrated into the model's architecture, enhancing the
model's ability to accurately represent and solve problems with inherent
periodic structures. We use a
uniform grid of size $\text{2001} \times \text{101}$ in the spatio-temporal
domain $[0,5] \times [0,2\pi]$ for creating initial, boundary, and residual
points. The PINN model is then trained using Adam optimizer for 100,000
iterations. Figure \ref{fig:convection_pinn}-(a) displays the PINN model's
performance and shows that the PINN model fails
to provide an accurate solution for this example. Specifically, the model predicts
the solution to be near zero after $t=1$. 
Figure \ref{fig:convection_pinn}-(b) depicts loss convergence of the
initial condition loss $\mathcal{L}^{\text{PINN}}_{ic}$ and the PDE residual
loss $\mathcal{L}^{\text{PINN}}_{res}$ during training.
Notably, the residual loss function decreases during
the initial few thousand training epochs and then shows negligible changes
throughout the remaining training period. This behavior indicates that the
predicted solution is biased towards an inappropriate local minimum.
The blue solid line and green dashed line in Figure
\ref{fig:convection_pinn}-(c) represent the residual loss of PINN at $t=0$ and
1, respectively, defined by
\begin{equation}
\label{eqn:pinn_residual_t}
\widetilde{\mathcal{L}}^{\text{PINN}}_{res}(t,\theta) = \frac{1}{N_x}\sum_{j=1}^{N_x}|U_t(t,x^j;\theta) + \mathcal{N}[U](t,x^j;\theta)|^2
\end{equation}
where $\mathcal{N}[U]=15 U_x$ and $x^j$'s are randomly sampled points on $\Omega$. As
$\widetilde{\mathcal{L}}^{\text{PINN}}_{res}(t,\theta)$ approximates the norm of
residual term at time $t$, we can use it to assess whether the approximate solution at
time $t$ satisfies the governing equation. 
A crucial observation is that
$\widetilde{\mathcal{L}}^{\text{PINN}}_{res}(1,\theta)$ decreases rapidly in the
initial training epochs, while
$\widetilde{\mathcal{L}}^{\text{PINN}}_{res}(0,\theta)$ remains large. 
 This reveals a potential violation of temporal causality, wherein the
solution at a later time step ($t=1$) is being optimized before adequate
training has taken place at an earlier time step ($t=0$).
Such a phenomenon can lead to a series of predictions at later time steps that,
while adhering to the governing equations, are biased towards an undesired local
minimum, in this case zero, which does not represent a physically viable
solution. This observation underscores the importance of adhering to temporal
causality in the optimization process; it is critical to prioritize the
minimization of residual loss at earlier time steps before moving onto later
ones, ensuring a more accurate and physically coherent solution.

\begin{remark}
The average of $\widetilde{\mathcal{L}}^{\text{PINN}}_{res}(t,\theta)$ evaluated on sampled points $t=t_0,t_1,\cdots,t_N \in [0,T]$ is equivalent to $\mathcal{L}^{\text{PINN}}_{res}(\theta)$ by the following.

\begin{equation*}
\begin{aligned}
\frac{1}{N}\sum_{n=0}^N\widetilde{\mathcal{L}}^{\text{PINN}}_{res}(t_n,\theta) &\approx \frac{1}{T}\int_0^T \widetilde{\mathcal{L}}^{\text{PINN}}_{res}(t,\theta)dt \approx \frac{1}{T}\frac{1}{|\Omega|}\int_0^T \int_\Omega |U_t(t,x;\theta) + \mathcal{N}[U](t,x;\theta)|^2 dxdt \\
&\approx \frac{1}{N_r}\sum_{j=1}^{N_r}|U_t(t_r^j,x_r^j;\theta) + \mathcal{N}[U](t_r^j,x_r^j;\theta)|^2 = \mathcal{L}^{\text{PINN}}_{res}(\theta)
\end{aligned}
\end{equation*}
\end{remark}

Unfortunately, this issue is not uncommon when applying PINNs to time-dependent
PDEs, as evidenced by a growing body of recent research 
\cite{mattey2022novel,wight2020solving,wu2022physics,wang2022long,wang2022respecting,krishnapriyan2021characterizing,du2021evolutional}.
In these cases, the accuracy of the solution deteriorate as the time variable increases,
leading to increasingly inaccurate predictions as time progresses. This becomes
particularly problematic in systems where the target solution exhibits a
substantial dependency on the initial conditions, such as those found in chaotic
systems. Therefore, it is crucial to develop methods that adequately account for
temporal causality. By doing so, we can ensure consistently accurate
predictions over the entire time domain.

%--------------------------------------------------------------
%
% SECTION: CEENs
%
%--------------------------------------------------------------
\section{Causality-enforced evolutional networks (CEENs)}\label{sec:tren}
In this section, we propose causality-enforced evolutional networks to
enforce the sequential learning and its parallelization to enhance the
computational efficiency, followed by a detailed analysis of the proposed method.

\subsection{Methodology of CEENs}\label{sec:methodology}
We present a method for solving time-dependent PDEs
that incorporates temporal causality into its formulation. We call this method
the causality-enforced evolutional network (\AL). As highlighted
in many studies, integrating the concept of temporal causality into model
construction when solving time-dependent PDEs is critical. Our proposed model
addresses this requirement by assigning distinct neural networks at each time
step and sequentially training them, beginning from the initial time step.

More specifically, we divide the temporal domain $[0,T]$ into $N$ equidistant
subintervals $0=t_0<t_1<\cdots<t_N=T$. At each time step $t=t_n$, we
assign a neural network model $U^n(x;\theta^n)$ with its unique parameter
$\theta^n$ to approximate $u(t_n,x)$. The loss function used to train the
designated neural network $U^n(x;\theta^n)$ is derived from the governing physical
law within the subinterval $[t_{n-1},t_n]$. 
To provide a detailed account, consider the integral form of the governing
equation \eqref{eqn:pde} over a subinterval $[t_{n-1},t_n]$:
\begin{equation}
\label{eqn:alternative_governing_equation}
u(t_n,x) = u(t_{n-1},x) - \int_{t_{n-1}}^{t_n}\mathcal{N}[u](t,x)dt.
\end{equation}
Linear multi-step methods in the classical numerical methods depend on the
choice of the approximation of the integral in~\eqref{eqn:alternative_governing_equation}.
For better illustration, from now on, we assume to use the trapezoidal rule to approximate
the integral:
\begin{equation}
\label{eqn:integral_approximation}
\int_{t_{n-1}}^{t_n}\mathcal{N}[u](t,x)dt = \frac{1}{2}h\big(\mathcal{N}[u](t_{n-1},x) + \mathcal{N}[u](t_n,x)\big) + \frac{\mathcal{N}[u]_{tt}(t_\ast,x)}{12}h^3
\end{equation}
where $t_\ast \in [t_{n-1},t_n]$ and $h=T/N$. 
Combining equation \eqref{eqn:integral_approximation} with the initial and
boundary conditions (\eqref{eqn:ic} and \eqref{eqn:bc}), we derive the
relationships between the neural networks for $n=1,2,\cdots,N$ as follows:
\begin{equation}
\label{eqn:relation_of_assigned_networks}
\begin{aligned}
U^0(x;\theta^0) &= g(x)\;\;\text{for $x \in \Omega$,} \\
\mathcal{B}[U^n](x;\theta^n) &= 0\;\;\text{for $x \in \partial\Omega$.} \\
U^n(x;\theta^n) &= U^{n-1}(x;\theta^{n-1}) - \frac{1}{2}h\big(\mathcal{N}[U^{n-1}](x;\theta^{n-1}) + \mathcal{N}[U^n](x;\theta^n)\big)\;\;\text{for $x \in \Omega$,}
\end{aligned}
\end{equation}
The loss function of \AL~ is then designed to ensure that the designated networks
satisfy \eqref{eqn:relation_of_assigned_networks} at each time step $t_n,
n=1,2,\cdots,N$:
\begin{equation*}
\mathcal{L}^{\ALtxt}(\theta) = \mathcal{L}^{\ALtxt}_{ic}(\theta) + \mathcal{L}^{\ALtxt}_{bc}(\theta) + \mathcal{L}^{\ALtxt}_{res}(\theta)
\end{equation*}
where the components of the loss function are defined as:
\begin{equation}
\label{eqn:tren_losses}
\begin{aligned}
\mathcal{L}^{\ALtxt}_{ic}(\theta) &= \frac{1}{N_x}\sum_{j=1}^{N_x}|U^0(x^j;\theta^0) - g(x^j)|^2, \\
\mathcal{L}^{\ALtxt}_{bc}(\theta) &= \frac{1}{N}\sum_{n=1}^N\frac{1}{N_b}\sum_{j=1}^{N_b}\mathcal{B}[U^n](x^j_{bc},\theta^n)^2=: \frac{1}{N}\sum_{n=1}^N\widetilde{\mathcal{L}}^{\ALtxt}_{bc}(t_n,\theta^n), \\
\mathcal{L}^{\ALtxt}_{res}(\theta) &= \frac{1}{N}\sum_{n=1}^N\frac{1}{N_x}\sum_{j=1}^{N_x}\Big|\frac{U^n(x^j;\theta^n) - U^{n-1}(x^j;\theta^{n-1})}{h} + \frac{1}{2}\big(\mathcal{N}[U^{n-1}](x^j;\theta^{n-1}) + \mathcal{N}[U^n](x^j;\theta^n)\big)\Big|^2 \\
&=:
    \frac{1}{N}\sum_{n=1}^N\widetilde{\mathcal{L}}^{\ALtxt}_{res}(t_n,\theta^n,\theta^{n-1}).
\end{aligned}
\end{equation}
$\{x^j\}_{j=1}^{N_x}$ and $\{x^j_{bc}\}_{j=1}^{N_{bc}}$ denote sampled
points to approximate $L^2$ norm on $\Omega$ and $\partial\Omega$, respectively.
The loss functions $\widetilde{\mathcal{L}}^{\ALtxt}_{bc}(t,\theta^n)$ and
$\widetilde{\mathcal{L}}^{\ALtxt}_{res}(t,\theta^n,\theta^{n-1})$ in
\eqref{eqn:tren_losses} represent the boundary and residual losses of \AL \  at
each time step $t=t_n, n=1,2,\cdots, N$.

In contrast to PINN, our method does not seek to
minimize the total loss $\mathcal{L}^{\ALtxt}(\theta)$ over the entire time
domain. Instead, to account for temporal causality, we prioritize sequential network
training, beginning from the initial time step. Accordingly, we split the loss
function $\mathcal{L}^{\ALtxt}(\theta)$ into:
\begin{equation*}
\mathcal{L}^\ALtxt(\theta) = \mathcal{L}^{\ALtxt}_0(\theta^0) + \frac{1}{N}\sum_{n=1}^N \mathcal{L}^{\ALtxt}_n(\theta^n,\theta^{n-1})
\end{equation*}
where
\begin{equation*}
\begin{aligned}
\mathcal{L}^{\ALtxt}_0(\theta^0) &= \mathcal{L}^{\ALtxt}_{ic}(\theta) \\
\mathcal{L}^{\ALtxt}_n(\theta^n,\theta^{n-1}) &=
    \widetilde{\mathcal{L}}^{\ALtxt}_{bc}(t_n,\theta^n) +
    \widetilde{\mathcal{L}}^{\ALtxt}_{res}(t_n,\theta^n,\theta^{n-1}).
\end{aligned}
\end{equation*}

We can enforce the sequential network training by systematically
minimizing the loss function $\mathcal{L}^{\ALtxt}_n$ in ascending order of
$n=0,1,\cdots,N$. More specifically, we initiate by training $\theta^0$ to
decrease $\mathcal{L}^{\ALtxt}_0(\theta^0)$ until it drops below a specified
tolerance, $\epsilon>0$. Once we identify the optimal parameter
$\overline{\theta^0}$ satisfying the condition
$\mathcal{L}^{\ALtxt}_0(\overline{\theta^0})<\epsilon$, we stop training 
$\theta^0$. We move on to train 
$\theta^1$ to minimize $\mathcal{L}^{\ALtxt}_1(\theta^1,\overline{\theta^0})$
until we identify $\overline{\theta^1}$ that satisfies the condition
$\mathcal{L}^{\ALtxt}_1(\overline{\theta^1},\overline{\theta^0})<\epsilon$.
Note that the value of $\overline{\theta^0}$ remains constant while
$\theta^1$ is being trained, implying that the past state $U^0$ is unaffected
by the current state $U^1$.
After identifying the optimal parameter $\overline{\theta^1}$, we proceed to
find $\overline{\theta^2}$ such that
$\mathcal{L}^{\ALtxt}_2(\overline{\theta^2},\overline{\theta^1})<\epsilon$. We
continue this iterative process until we obtain the final trained parameter
$\overline{\theta^N}$ satisfying
$\mathcal{L}^{\ALtxt}_N(\overline{\theta^N},\overline{\theta^{N-1}})<\epsilon$.
This training procedure sequentially reduces the residual loss at each time
step, unlike PINN, thereby ensuring that the solution respects 
the temporal causality.

%--------------------------------------------------------------
% SUB_SECTION: Parallelization of CEENs
%--------------------------------------------------------------
\subsection{Enhancing CEENs with Parallelization}\label{sec:parallelization}

The total number of neural networks required for training is $N+1=T/h+1$, which
can be quite large for a long time $T$ or small time step size $h$.
While training each neural network over a short time interval is
generally quicker than training the original PINN over the whose time domain,
the process could still demand considerable computational resources. To
mitigate these computational demands, we propose a parallelization algorithm.
 This approach enables us to simultaneously train $M$
successive neural networks instead of adhering to the traditional method of
training a single neural network for each time step.
To be concrete, in the beginning of the training
process, we train the networks $U^0,U^1,\cdots,U^{M-1}$ instead of solely
training $U^0$. In this case, the loss function
$\mathcal{L}^{\ALtxt}_{0:M-1}$ of their parameters
$\theta^0,\theta^1,\cdots,\theta^{M-1}$ is the sum of
$\mathcal{L}^{\ALtxt}_0,\mathcal{L}^{\ALtxt}_1,\cdots,\mathcal{L}^{\ALtxt}_{M-1}$,
i.e.,
\begin{equation*}
\mathcal{L}^{\ALtxt}_{0:M-1}(\theta^0,\theta^1,\cdots,\theta^{M-1}) = \mathcal{L}^{\ALtxt}_0(\theta^0) + \sum_{k=1}^{M-1}\mathcal{L}^{\ALtxt}_k(\theta^k,\theta^{k-1}).
\end{equation*}
In order to use gradient-based optimization methods, we require its gradient:
\begin{equation*}
\nabla_{(\theta^0,\theta^1,\cdots,\theta^{M-1})} \mathcal{L}^{\ALtxt}_{0:M-1} = \nabla_{\theta^0}\mathcal{L}^{\ALtxt}_0 + \sum_{k=1}^{M-1}\nabla_{(\theta^k,\theta^{k-1})}\mathcal{L}^{\ALtxt}_k,
\end{equation*}
which can be calculated in parallel by concurrently computing each gradient term across
multiple GPU cores. As a result, the computational cost of training multiple networks
becomes comparable to that of training a single network, with a decrease in the
total number of required epochs, consequently reducing the computational cost.

It is important to note that we must account for temporal causality during
parallel training. For example, should
$\mathcal{L}^{\ALtxt}_1(\theta^1,\theta^0)<\epsilon$ be achieved before
$\mathcal{L}^{\ALtxt}_0(\theta^0)<\epsilon$, it would be incorrect to stop the
training of $\theta^1$ before the completion of $\theta^0$ training, as this
would violate temporal causality. Therefore, we initially optimize the loss
$\mathcal{L}^{\ALtxt}_{0:M-1}$ until $\mathcal{L}^{\ALtxt}_0(\theta^0)$ is
sufficiently reduced. Once we find
$\overline{\theta^0}$ that makes $\mathcal{L}^{\ALtxt}_0(\overline{\theta^0}) <
\epsilon$, we stop training for the parameter $\theta^0$ and focus on minimizing
the subsequent loss function $\mathcal{L}^{\ALtxt}_{1:M-1}$:
\begin{equation*}
\mathcal{L}^{\ALtxt}_{1:M-1}(\theta^1,\theta^2,\cdots,\theta^{M-1}) = \mathcal{L}^{\ALtxt}_1(\theta^1,\overline{\theta^0}) + \sum_{k=2}^{M-1}\mathcal{L}^{\ALtxt}_k(\theta^k,\theta^{k-1}).
\end{equation*}
Upon finding $\overline{\theta^1}$ satisfying
$\mathcal{L}^{\ALtxt}_1(\overline{\theta^1},\overline{\theta^0}) < \epsilon$,
we stop the training of $\theta^1$, and then proceed to minimize the loss function
$\mathcal{L}^{\ALtxt}_{2:M-1}$, which comprises the remaining loss functions. 
This iterative process continues until we find $\overline{\theta^{M-1}}$ satisfying
$\mathcal{L}^{\ALtxt}_{M-1}(\overline{\theta^{M-1}},\overline{\theta^{M-2}})<\epsilon$. 
The training of the next set of neural networks of size $M$ then
follows the same pattern.

\begin{remark}
As described in the above parallelization strategy, it is crucial
to maintain a sequential order even when training $M$ networks
simultaneously. This constraint introduces additional overhead, explaining
why the computational cost doesn't scale linearly as $M$ increases as
 demonstrated in Section~\ref{sec:numerical}.
A large $M$ may be constrained by the GPU resources.
 Consequently, the optimal selection of $M$ must be determined
heuristically, considering the available GPU capacity.
\end{remark}

%--------------------------------------------------------------
% SUB_SECTION: Analysis of CEENs
%--------------------------------------------------------------
\subsection{Analysis on CEENs}\label{sec:analysis}
In this subsection, we discuss a detailed analysis of the proposed method such
as the error bound of the solution at each subinterval, the training of initial
condition, memory efficiency, and comparison with the conventional numerical
methods.
\subparagraph{The choice of time step size and tolerance.}
The time step size $h$ and the tolerance $\epsilon$ are important
hyperparameters for the accuracy of the prediction, with the potential to
introduce significant errors at each time step if chosen incorrectly.
Specifically, a large time step size $h$ can result in considerable
approximation errors, which are proportional to $h^2$ due to the
trapezoidal rule. Conversely, setting the tolerance $\epsilon$ too high may lead
to substantial errors in the predicted solution stemming from insufficient
training. Note that the exact solution over a subinterval $[t_{n-1},t_n]$ should
satisfy the following equation, which can be derived from \eqref{eqn:alternative_governing_equation} and
\eqref{eqn:integral_approximation},
\begin{equation*}
    u(t_n,x) - u(t_{n-1},x) + \frac{1}{2}h(\mathcal{N}[u](t_{n-1},x) +
    \mathcal{N}[u](t_n,x) +
    \frac{\mathcal{N}[u]_{tt}(t_\ast,\cdot)}{12}h^3 =0.
\end{equation*}
Since $U^n(x;\theta^n)$ with parameters $\theta^n$ represents the solution 
 at $t_n$ satisfying the equation~\eqref{eqn:relation_of_assigned_networks}, we can obtain the
following $L^2$ error:
\begin{equation}
\label{eqn:error_at_each_time_step}
\begin{aligned}
&\;\;\|U^n - U^{n-1} + \frac{1}{2}h(\mathcal{N}[U^{n-1}] + \mathcal{N}[U^n]) + \frac{\mathcal{N}[u]_{tt}(t_\ast,\cdot)}{12}h^3\|_{L^2(\Omega)} \\
&\leq \|U^n - U^{n-1} + \frac{1}{2}h(\mathcal{N}[U^{n-1}] + \mathcal{N}[U^n])\|_{L^2(\Omega)} + \frac{1}{12}h^3\|\mathcal{N}[u]_{tt}(t_\ast,\cdot)\|_{L^2(\Omega)} \\
&\approx \Big[\frac{|\Omega|}{N_x}\sum_{i=1}^{N_x}\Big|U^n(x^i;\theta^n) - U^{n-1}(x^i;\theta^{n-1}) + \frac{1}{2}h\big(\mathcal{N}[U^{n-1}](x^i;\theta^{n-1}) + \mathcal{N}[U^n](x^i;\theta^n)\big)\Big|^2\Big]^{1/2} \\
&\;\;\;\;+ \frac{1}{12}h^3\|\mathcal{N}[u]_{tt}(t_\ast,\cdot)\|_{L^2(\Omega)}\\
&\leq |\Omega|^{1/2}(\mathcal{L}_n^{\ALtxt})^{1/2}h + \frac{1}{12}h^3\|\mathcal{N}[u]_{tt}(t_\ast,\cdot)\|_{L^2(\Omega)} \\
&\leq |\Omega|^{1/2}\epsilon^{1/2}h + \frac{1}{12}h^3\|\mathcal{N}[u]_{tt}(t_\ast,\cdot)\|_{L^2(\Omega)}=:e_1(\epsilon,h) + e_2(h)
\end{aligned}
\end{equation}
where $e_1(\epsilon,h) := |\Omega|^{1/2}\epsilon^{1/2}h$ and $e_2(h) :=
\frac{1}{12}h^3\|\mathcal{N}[u]_{tt}(t_\ast,\cdot)\|_{L^2(\Omega)}$.
Note that the error over a subinterval $[t_{n-1},t_n]$ consists of two terms,
$e_1(\epsilon,h)$ and $e_2(h)$. $e_1(\epsilon,h)$ represents the training loss,
dependent on $\epsilon$ and $h$, while $e_2(h)$ accounts for the consistency
error of the trapezoidal rule, dependent on $h$.
A large value of $\epsilon$ can lead to a significant $e_1(\epsilon,h)$ error at each
time step. Conversely, choosing a small $\epsilon$
may not be advantageous either. Despite a small $\epsilon$ leading to a lower
value of $e_1$, it could entail an increased training cost to bring
$\mathcal{L}^{\ALtxt}_n$ below $\epsilon$ at each time step. Given that the
error $e_2$ is inevitably incurred by the trapezoidal approximation, choosing an
$\epsilon$ that significantly reduces $e_1$ below $e_2$ is unnecessary.
Ideally, $\epsilon$ should be selected to satisfy $e_1(\epsilon,h) \approx
e_2(h)$. However, this principle might not always be practical as the value of
$\|\mathcal{N}[u]_{tt}(t_\ast,\cdot)\|_{L^2(\Omega)}$ is generally unknown.
Regardless, this principle can still offer valuable guidance for choosing $\epsilon$
as the time step size $h$ varies.

The error bound, characterized by two components $e_1$ and $e_2$, reveals the
order of the accuracy with respect to the time step size and tolerance.
If $\epsilon$ is sufficiently small such that $e_1$'s impact on total
error is minor, then the error is dominated by $e_2$. This
suggests a second-order accuracy with respect to $h$, consistent with
expectations from the Trapezoidal rule. Conversely, when $h$ is reduced to a
point where $e_2$'s effect on the error is minimal, $e_1$ becomes the main
contributor to the error, indicating a half-order accuracy in terms of
$\epsilon$. The validation of this relationship will be examined in
Section~\ref{sec:convection}, providing empirical support for these theoretical
assertions.

\subparagraph{The training of initial condition.} 
One crucial aspect of the method that requires careful consideration is the
training of the initial condition \eqref{eqn:ic}. Note that a small value of
$\mathcal{L}^{\ALtxt}_0$ does not automatically ensure an accurate approximation of
$\mathcal{N}[g]$ by $\mathcal{N}[U^0]$, which appears in
$\mathcal{L}^{\ALtxt}_1$. This is primarily because the differential operator
$\mathcal{N}$ is generally not a continuous linear operator.
In other words, the following inequality with some constant $C>0$ does not hold
in general:
\begin{equation*}
\|\mathcal{N}[U^0] - \mathcal{N}[g]\|_{L^2(\Omega)} \leq C\|U^0 - g\|_{L^2(\Omega)}
\end{equation*}
This can lead to significant errors when calculating
$\mathcal{L}^{\ALtxt}_1$ as $\mathcal{N}[U^0]$ may deviate from $\mathcal{N}[g]$.
The initial condition loss and residual loss at $t=0$ in
Figure~\ref{fig:convection_pinn}-(b,c) effectively capture this phenomenon.
To address this issue, we propose to utilize the modified loss of
$\mathcal{L}^{\ALtxt}_0$ incorporating the derivatives:
\begin{equation} \label{eqn:loss_ic}
\mathcal{L}^{\ALtxt}_0(\theta) = \sum_{k=0}^K\frac{1}{N_x}\sum_{j=1}^{N_x}|D^kU^0(x^j;\theta^0) - D^kg(x^j)|^2,
\end{equation}
where $D^k u$ denotes the list of all $k$-th order partial derivatives of $u$
and $\mathcal{N}$ is assumed to be the Lipschitz function of $u$, $Du$,
$D^2u$,$\cdots$,$D^Ku$. This adjustment ensures that a small value of
$\mathcal{L}^{\ALtxt}_0$ corresponds to a more accurate approximation of $\mathcal{N}[g]$
by $\mathcal{N}[U^0]$. For implementing this formulation, we assume that precise
knowledge of the initial conditions or data on their derivatives is available.

\subparagraph{Memory efficiency.} 
The memory efficiency of \AL s is superior compared to PINNs, attributed to
several reasons. Firstly, unlike PINNs that aim to model solutions across the
entire spatio-temporal domain, \AL s focus on modeling solutions within a
short temporal domain, thus leveraging smaller networks. This
implies that the number of parameters to be updated during the training of \AL~is
significantly lower compared to PINNs. Secondly, while PINNs'
training utilizes points spanning the entire spatio-temporal domain, \AL s
employ fewer training points, confined to a short temporal domain. Additionally,
\AL s avoid the usage of automatic differentiation in the time direction,
resulting in a further decrease in memory usage. 

Therefore, \AL~demands considerably less memory for gradient computation in
updating parameters compared to PINN. While PINN occupies the GPU with the
update of numerous parameters, \AL~processes parameter updates swiftly, even
within constrained GPU environments.
This memory efficiency potentially makes large-scale simulations of complex
problems feasible, which might prove challenging or even unachievable when using
PINNs.

\subparagraph{Maximum training epoch.} 
There may be instances when $\mathcal{L}^{\ALtxt}_n$ cannot be minimized to the
desired extent.
To this end, we introduce the maximum number of training epochs, denoted
by $E$, to prevent excessive iterations so that the training of $\theta^n$ is
finished even if $\mathcal{L}^{\ALtxt}_n<\epsilon$ is not achieved. If the loss
is not sufficiently reduced over many consecutive time steps, errors can
accumulate and negatively affect the accuracy of the overall solution. In this
case, we should consider a change in the architecture of the neural network as
it may not be suitable to represent the solution. For example, for modeling
high-frequency solutions, we can consider using a Fourier feature network
\cite{tancik2020fourier,wang2021eigenvector}.

\subparagraph{Comparison with classical numerical methods}

The proposed method is similar to classical numerical methods for solving
time-dependent PDEs in the sense that it sequentially computes the solution at each time
step. However, CEEN presents several advantages over these classical methods.
Firstly, CEEN can precisely compute the spatial derivatives of the solution. In
contrast, conventional mesh-based numerical methods often require the dense
placement of spatial points to achieve the desired accuracy in spatial
derivatives. CEEN, by leveraging a neural network for each time step,
facilitates the exact computation of derivatives via automatic differentiation,
thereby eliminating approximation errors common in traditional methods.
Secondly, CEEN offers flexibility in the positioning of spatial points.
Traditional numerical methods might necessitate specific arrangements of
spatial points to ensure accurate computation of derivatives, a restriction CEEN
does not impose. This adaptability proves advantageous, particularly in
scenarios where data is only available at certain locations within the spatial
domain.
Thirdly, after training on specific spatial points, the solution at any
arbitrary point in CEEN can be instantly obtained with a single forward pass
through the trained network. This feature significantly enhances the efficiency and
applicability of the method, providing immediate solutions at any desired
location.
Finally, leveraging automatic differentiation for the precise calculation of
spatial derivatives enables us to overcome traditional constraints on time step
and spatial grid sizes mandated by classical methods for stability, such as the
Courant-Friedrichs-Lewy (CFL) condition. This advantage allows for the selection
of time step sizes independent of grid dimensions. However, unlike classical
methods that have established convergence analyses, such as the Lax Equivalence
Theorem, an equivalent convergence analysis for PINNs, addressing aspects like
generalization error, is not yet available. This area, requiring a more profound
understanding of PINNs, remains beyond the scope of this paper but represents a
significant avenue for future research.

\begin{figure}[htbp]
    \centering
    \includegraphics[width=0.77\textwidth]{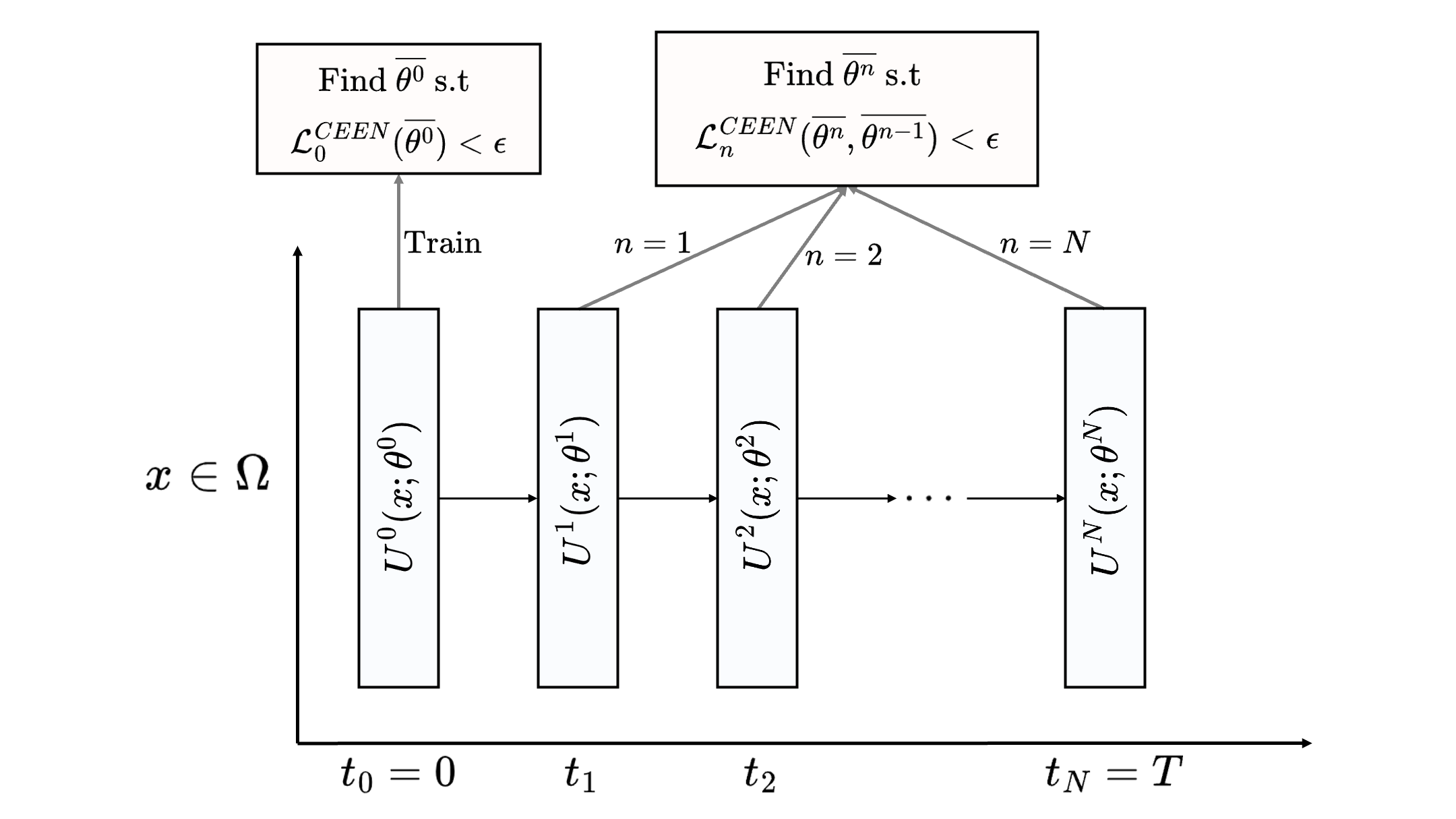}
    \caption{Schematic of \AL. The temporal domain is discretized into
    subintervals, with a neural network assigned to each time step. The solution
    is obtained by sequentially training each network, beginning from the
    initial time. This approach reflects temporal causality in the training process.}
    \label{fig:tren}
\end{figure}

\subparagraph{} 
A schematic diagram representing \AL~is shown in Figure \ref{fig:tren}. Algorithm \ref{alg:tren} summarizes the proposed method outlined above.

\begin{algorithm}[htbp]
    \caption{Causality-enforced evolutional network (\AL)}
    \textbf{Input:} Maximum epoch $E$, Number of networks computed in parallel $M$, tolerance $\epsilon$, learning rate $\eta$.
    \begin{algorithmic}[1]
    \State Initialize $\theta=(\theta^0,\theta^1,\cdots,\theta^N)$ and $m=0$
    \Repeat
    \For{$n=m,m+1,\cdots,m+M-1$}
        \State{$\mathcal{L}^{\ALtxt}_{n:m+M-1}(\theta^n,\theta^{n+1},\cdots,\theta^{m+M-1})=\mathcal{L}^{\ALtxt}_n(\theta^{n},\overline{\theta^{n-1}}) + \sum_{k=n+1}^{m+M-1}\mathcal{L}^{\ALtxt}_k(\theta^k,\theta^{k-1})$}
        \While{$\mathcal{L}^{\ALtxt}_n(\theta^n,\overline{\theta^{n-1}}) < \epsilon$}
            \State $(\theta^n,\theta^{n+1},\cdots,\theta^{n+M-1}) = (\theta^n,\theta^{n+1},\cdots,\theta^{n+M-1}) - \eta \nabla_{(\theta^n,\theta^{n+1},\cdots,\theta^{n+M-1})} \mathcal{L}^{\ALtxt}_{n:m+M-1}$
            \If {$\mathcal{L}^{\ALtxt}_n < \epsilon$ is not achieved for $E$ epochs}
                \State break
            \EndIf
        \EndWhile
        \State Save the parameter $\theta^n$ as $\overline{\theta^n}$
    \EndFor
    \State{$m = m+M$}
    \Until{All network parameters are trained}
    \end{algorithmic}
    \textbf{Output:} Trained parameters $\overline{\theta^0},\overline{\theta^1},\cdots,\overline{\theta^N}$
    \label{alg:tren}
\end{algorithm}

%--------------------------------------------------------------
%
% SECTION: NUMERICAL RESULTS
%
%--------------------------------------------------------------
\section{Numerical results}\label{sec:numerical}

We demonstrate the effectiveness of our proposed CEEN algorithm for solving a
range of time-dependent PDEs over long time intervals.  Specifically, we apply
our method to the Korteweg-De Vries (KdV), Schr{\"o}dinger, convection, and Allen-Cahn
equations. The performance of our method is then compared to that of the PINN.
As summarized in Table~\ref{tab:simulations}, our algorithm (\AL) consistently
delivers higher accuracy at the final time $t=5$ across all these problems
compared to PINN. This underscores the superior performance of our
proposed method.

\begin{table}
    \centering
    \caption{Relative errors of \AL~and PINN at the final time $t=5$ for four
    PDE problems.}
    \label{tab:simulations}    
    \begin{tabular}{l|r|r}
        \hline
        \multicolumn{1}{c|}{\multirow{2}{*}{Example}} &
          \multicolumn{2}{c}{Relative error} \\
          \cline{2-3}
        \multicolumn{1}{c|}{} &
          \multicolumn{1}{c|}{\AL} &
          \multicolumn{1}{c}{PINN} \\
          \hline
        KdV & 5.314e-3 & 1.237e-0 \\
        Schr{\"o}dinger & 4.793e-3 & 2.346e-1  \\
        Convection & 5.465e-3 & 1.000e-0 \\
        Allen-Cahn & 4.222e-2 & 1.394e-0 \\
        \hline
    \end{tabular}
\end{table}

We compute reference solutions utilizing the py-pde package
\cite{zwicker2020pypde}, with a time step size of 1e-7. 
The prediction of both \AL~and PINN are compared against these reference
solutions. Across all problems, parameters are initialized via the Glorot normal scheme, the
activation function is set as hyperbolic tangent, and the maximum training
epochs are capped at $E=\text{100,000}$. For each problem, the
value of $K$ in \eqref{eqn:loss_ic}, which is employed to learn the initial
condition, is set to be the highest order of the derivative present in the 
associated differential operator $\mathcal{N}$.
All networks are trained through the Adam optimizer with $\beta=(0.999,0.999)$, a configuration providing stable
learning \cite{ma2022qualitative}. For all problems except Allen-Cahn problem, we enforce periodic boundary conditions as
hard constraints by utilizing an appropriate Fourier features embedding of the input
as outlined in~\cite{dong2021method}.

%--------------------------------------------------------------
% SUBSECTION: KDV
%--------------------------------------------------------------
\subsection{KdV equation}

The first example aims to demonstrate the proposed method's capability to
address PDEs characterized by high order derivatives and dispersive nonlinearity.
For this purpose, we consider the
KdV equation, a mathematical model representing shallow water surface waves and ionic
acoustic waves in plasmas. The KdV equation takes the form
\begin{equation}
\label{eqn:kdv}
\begin{aligned}
u_t + uu_x + 0.0025u_{xxx}&=0,\;\;\text{for $t \in [0,5], x \in[-1,1]$} \\
u(0,x) &= \cos\pi x, \\
u(t,-1) &= u(t,1), \\
u_x(t,-1) &= u_x(t,1).
\end{aligned}
\end{equation}
To apply our method, we discretize the time domain with a
time step size $h=\text{1e-2}$. 
A fully-connected network with two hidden layers, each containing 16 neurons, is
utilized to model the solution $U^i(x;\theta^i)$ at each time step $t_i$. We
select $N_x=151$ equidistant points on the spatial domain and set the tolerance
to $\epsilon=\text{1e-4}$. The values of $K$ and $M$ used are 3
and 5, respectively.

For a fair comparison of our method with PINN, we design the architectures for
PINN so that both PINN and CEEN have nearly the same number of parameters.
Specifically, our method uses 501 neural
networks, each consisting of 465 parameters, resulting in a total of 232,965 parameters.
The PINN model, on the other hand, is constructed with a fully-connected network with two
hidden layers each containing 480 neurons. In total, the
PINN model has 237,121 parameters. 
\begin{figure}
    \centering
    \begin{subfigure}{\linewidth}
        \centering \includegraphics[height=4.5cm]{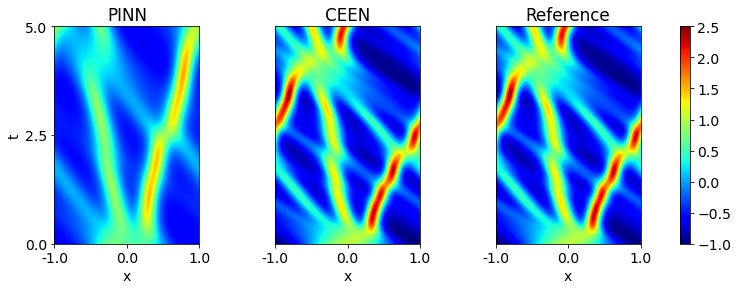}
        \caption{}
    \end{subfigure}
    \centering
    \begin{subfigure}{\linewidth}
        \centering \includegraphics[height=4.5cm]{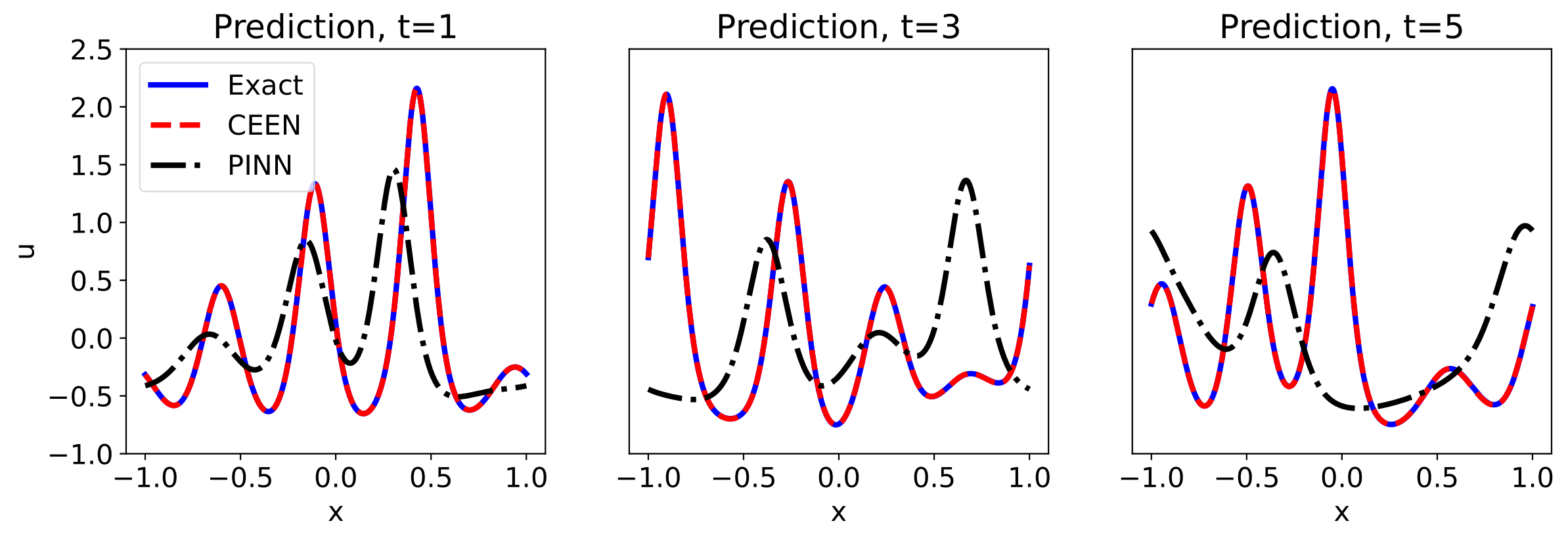}
        \caption{}
    \end{subfigure}
    \caption{\textit{KdV equation:} 
    Comparison of the prediction of \AL~and PINN against the reference
    solution. (a) displays the prediction across the entire spatio-temporal
    domain. (b) displays it at three different times: $t=1,3,5$. \AL~agrees well
    with the reference in the entire domain while PINN exhibits a less accurate
    approximation of the solution even at an earlier time $t=1$.}
    \label{fig:kdv_solution}
\end{figure}

\begin{figure}
    \centering
    \begin{subfigure}{0.49\linewidth}
        \centering \includegraphics[height=4.5cm]{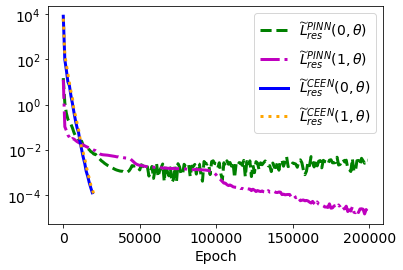}
        \caption{}
    \end{subfigure}
    \centering
    \begin{subfigure}{0.49\linewidth}
        \centering \includegraphics[height=4.5cm]{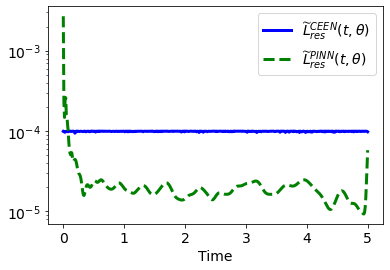}
        \caption{}
    \end{subfigure}
    \caption{Residual losses of \AL~and PINN for the KdV equation \eqref{eqn:kdv}. (a) Epoch vs residual loss (b) Time vs residual loss.}
    \label{fig:kdv_residual_loss}
\end{figure}

As shown in Figure \ref{fig:kdv_solution}, our proposed method accurately
predicts the solution, demonstrating its efficiency in solving long-time
integration of the PDE by respecting temporal causality. Conversely, PINN
fails to produce an accurate prediction even at earlier time, e.g., $t=1$ as
shown in Figure~\ref{fig:kdv_solution}-(b). The residual loss of PINN at $t=0$,
denoted as $\widetilde{\mathcal{L}}^{\text{PINN}}_{res}(0;\theta)$, stalls after
a certain epoch as displayed in Figure \ref{fig:kdv_residual_loss}-(a), while
that of PINN at $t=1$, $\widetilde{\mathcal{L}}^{\text{PINN}}_{res}(1;\theta)$, continues to decrease.
This suggests that PINN trains the solutions at later time steps even when the
solution at earlier time steps are insufficiently trained, violating temporal
causality. 
In contrast, \AL~decreases the residual losses,
$\widetilde{\mathcal{L}}^{\ALtxt}_{res}(t;\theta)$, to the tolerance
$\epsilon=\text{1e-4}$ at both $t=0$ and 1. Figure
\ref{fig:kdv_residual_loss}-(b) displays the residual losses of PINN and \AL~at
each time. Notably, despite PINN having low residual
loss across most of the time domain, it fails to predict a solution due to the
high loss in the initial time step. Our sequential learning strategy effectively
mitigates this issue by reducing the residual loss at each step sequentially
from the initial time step.

%--------------------------------------------------------------
% SUBSECTION: SCHRODINGER
%--------------------------------------------------------------
\subsection{Schr{\"o}dinger equation}

We next consider the Schr{\"o}dinger equation, which governs a nonlinear wave propagation in quantum mechanical
systems. It takes the form:
\begin{equation}
\label{eqn:schrodinger}
\begin{aligned}
i\psi_t + 0.5\psi_{xx} + |\psi|^2\psi&=0,\;\;\text{for $t \in [0,5], x \in[-5,5],$} \\
\psi(0,x) &= 2\,\text{sech}x, \\
\psi(t,-5) &= \psi(t,5), \\
\psi_x(t,-5) &= \psi_x(t,5).
\end{aligned}
\end{equation}
In fact, if $u$ and $v$ denote the real and imaginary parts of $\psi$, respectively, they satisfy the following coupled governing equations.
\begin{equation*}
\begin{aligned}
u_t + 0.5v_{xx} + v(u^2 + v^2) &=0, \\
v_t - 0.5u_{xx} - u(u^2 + v^2) &=0.
\end{aligned}
\end{equation*}

To apply our method, we discretize the time domain with the step size
$h=\text{1e-2}$. The solutions $u$ and $v$ are jointly represented at each time
step using a fully connected neural network with two outputs. The architecture
of the network includes two hidden layers, each containing 16 neurons. We use
$N_x=501$ equidistant points as sampled points on the
spatial domain and the tolerance is $\epsilon=\text{1e-5}$. 
The chosen values for K and M are 2 and 2, respectively.
The setting for PINN method is established in a similar way as outlined in the
previous examples. 

\begin{figure}
    \centering
    \begin{subfigure}{\linewidth}
        \centering \includegraphics[height=4.5cm]{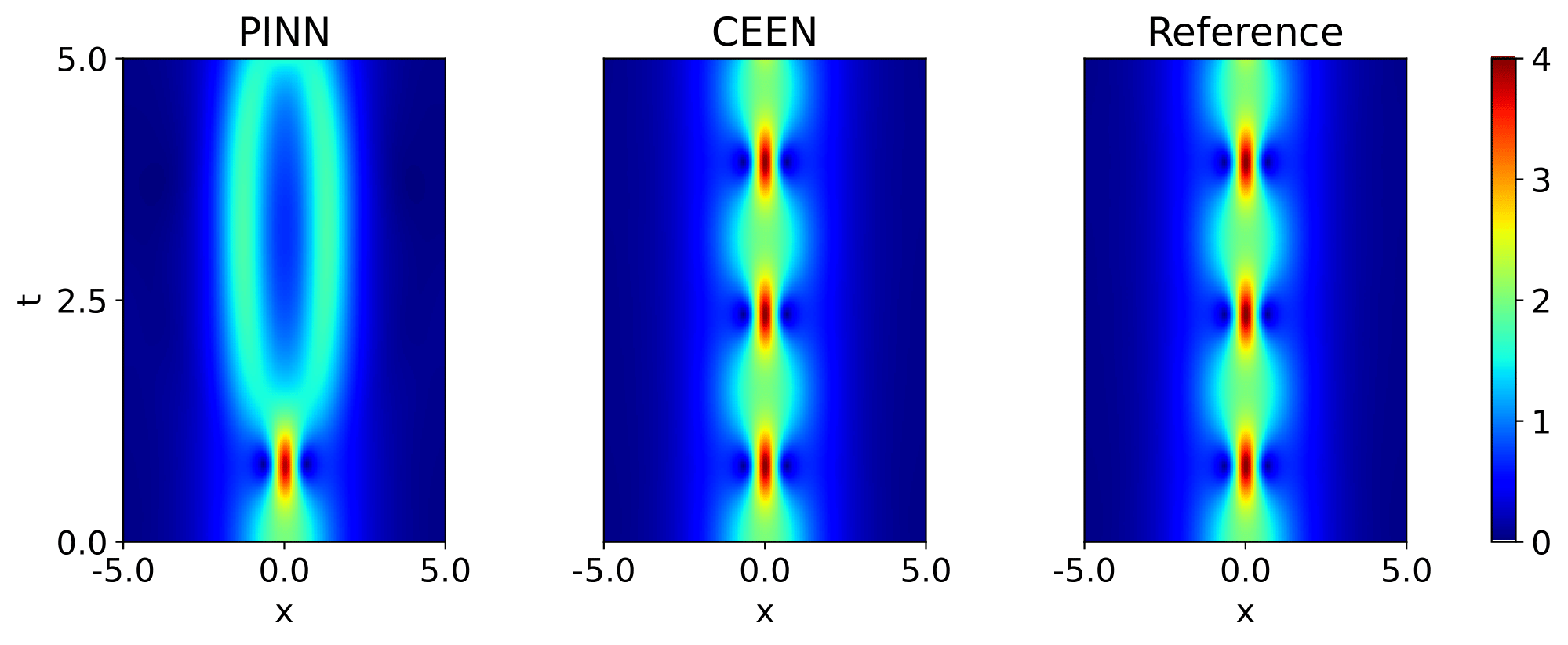}
        \caption{}
    \end{subfigure}
    \centering
    \begin{subfigure}{\linewidth}
        \centering \includegraphics[height=4.5cm]{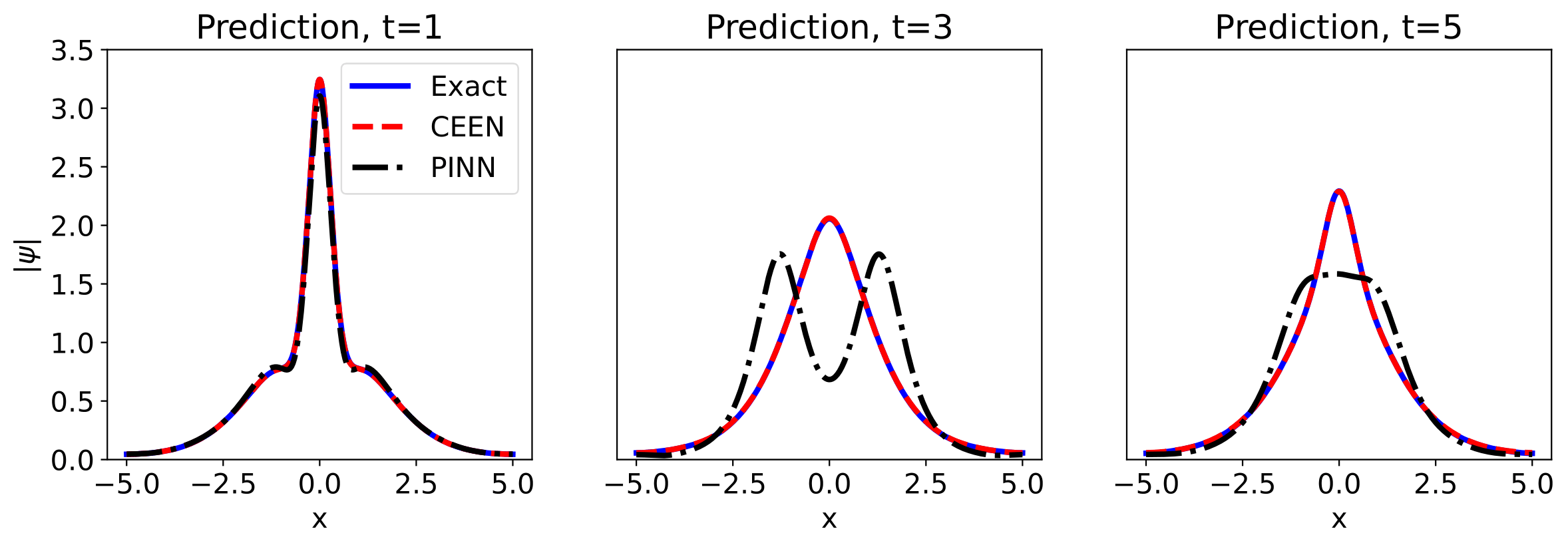}
        \caption{}
    \end{subfigure}
    \caption{\textit{Schr{\"o}dinger equation:} 
    Comparison of the prediction of \AL~and PINN against the reference
    solution. (a) displays the prediction across the entire spatio-temporal
    domain. (b) displays it at three different times: $t=1,3,5$. \AL~agrees well
    with the reference in the entire domain while PINN closely approximates the
    solution at an earlier time $t=1$. However, discrepancies between PINN
    predictions and the reference solution become apparent at later times $t=3$
    and $t=5$.}
    \label{fig:schrodinger_solution}
\end{figure}

\begin{figure}
    \centering
    \begin{subfigure}{0.49\linewidth}
        \centering \includegraphics[height=4.5cm]{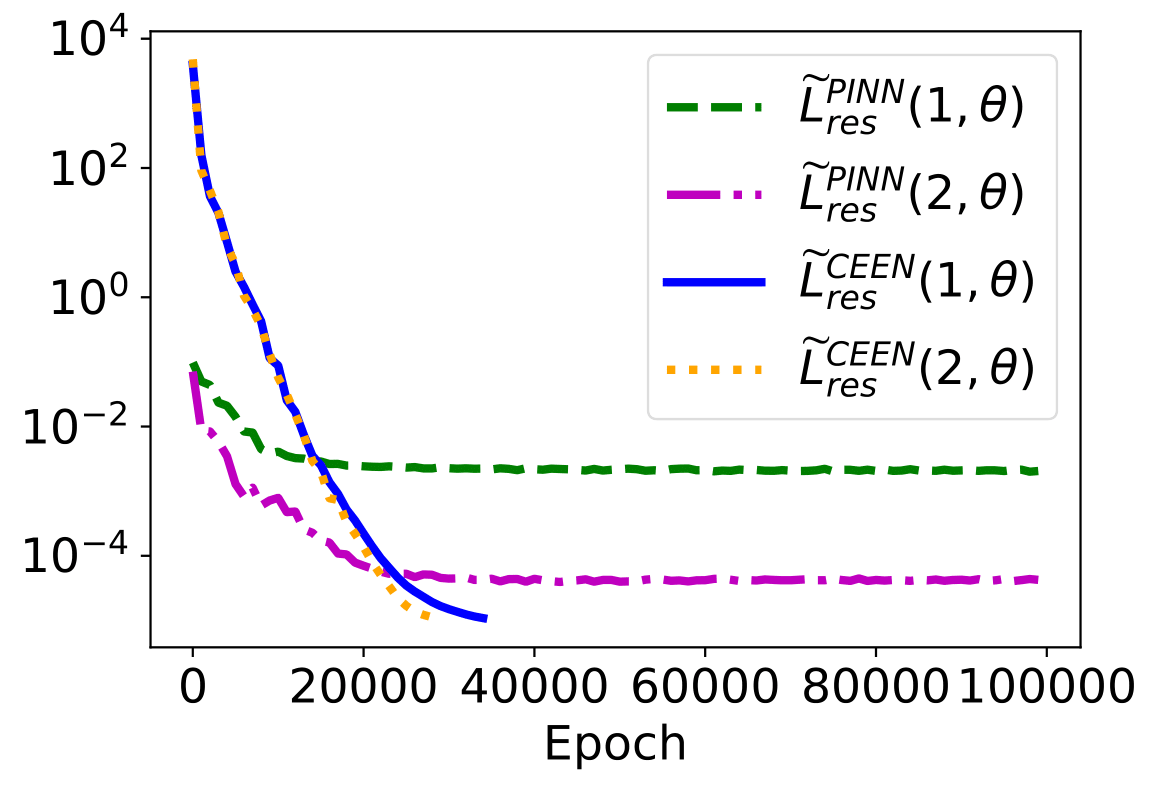}
        \caption{}
    \end{subfigure}
    \centering
    \begin{subfigure}{0.49\linewidth}
        \centering \includegraphics[height=4.5cm]{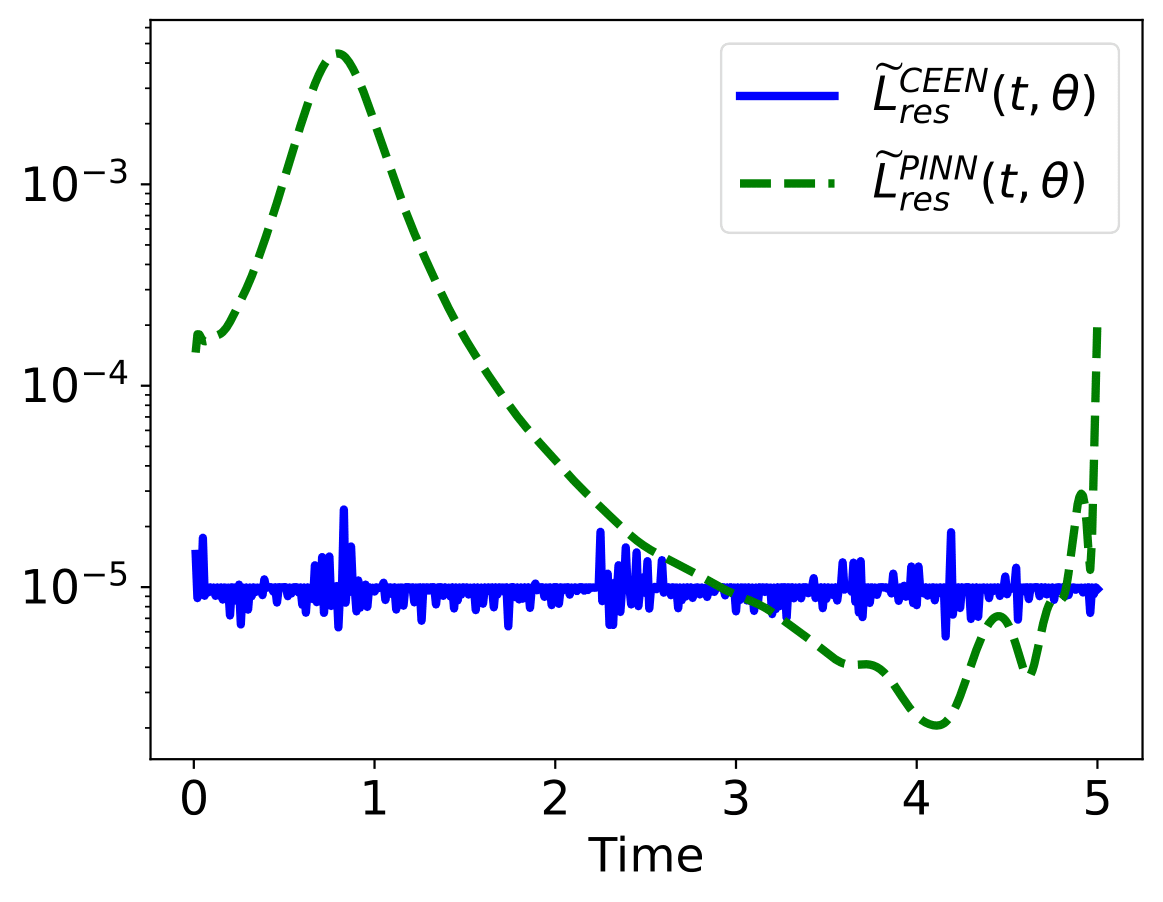}
        \caption{}
    \end{subfigure}
    \caption{Residual losses of \AL~and PINN for the Schr{\"o}dinger equation \eqref{eqn:schrodinger}. (a) Epoch vs residual loss (b) Time vs residual loss.}
    \label{fig:schrodinger_residual_loss}
\end{figure}

The results are shown in Figure \ref{fig:schrodinger_solution} where
our method shows good agreement with the reference solution across the whole
spatio-temporal domain. However, while PINN's predictions shows good agreement up to
around $t=1$, they start to deviate from the reference solution after $t=1$. 
Figure \ref{fig:schrodinger_residual_loss}-(a) shows that
$\widetilde{\mathcal{L}}^{\text{PINN}}_r(1;\theta)$ remains high while 
$\widetilde{\mathcal{L}}^{\text{PINN}}_r(2;\theta)$ is sufficiently reduced and
remains low during training. This indicates a violation of temporal causality,
as the solution at $t=2$ appears to be settled before the solution at $t=1$ has
been adequately trained. On the other hand, the residual losses of \AL~are sufficiently
reduced at both $t=1$ and 2 to the prescribed threshold $\epsilon=\text{1e-5}$. 
Figure \ref{fig:schrodinger_residual_loss}-(b) demonstrates that
\AL~achieves a satisfactory reduction of residual losses across the entire time
domain while the residual losses of PINN remain noticeably elevated near
$t=1$.

%--------------------------------------------------------------
% SUBSECTION: CONVECTION
%--------------------------------------------------------------
\subsection{Convection equation}\label{sec:convection}

\begin{figure}
    \centering
    \begin{subfigure}{\linewidth}
        \centering \includegraphics[height=4.5cm]{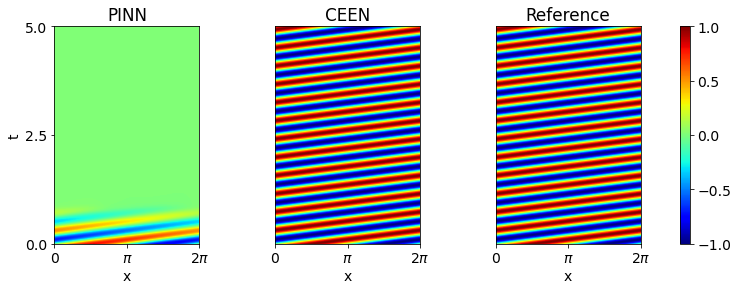}
        \caption{}
    \end{subfigure}
    \centering
    \begin{subfigure}{\linewidth}
        \centering \includegraphics[height=4.5cm]{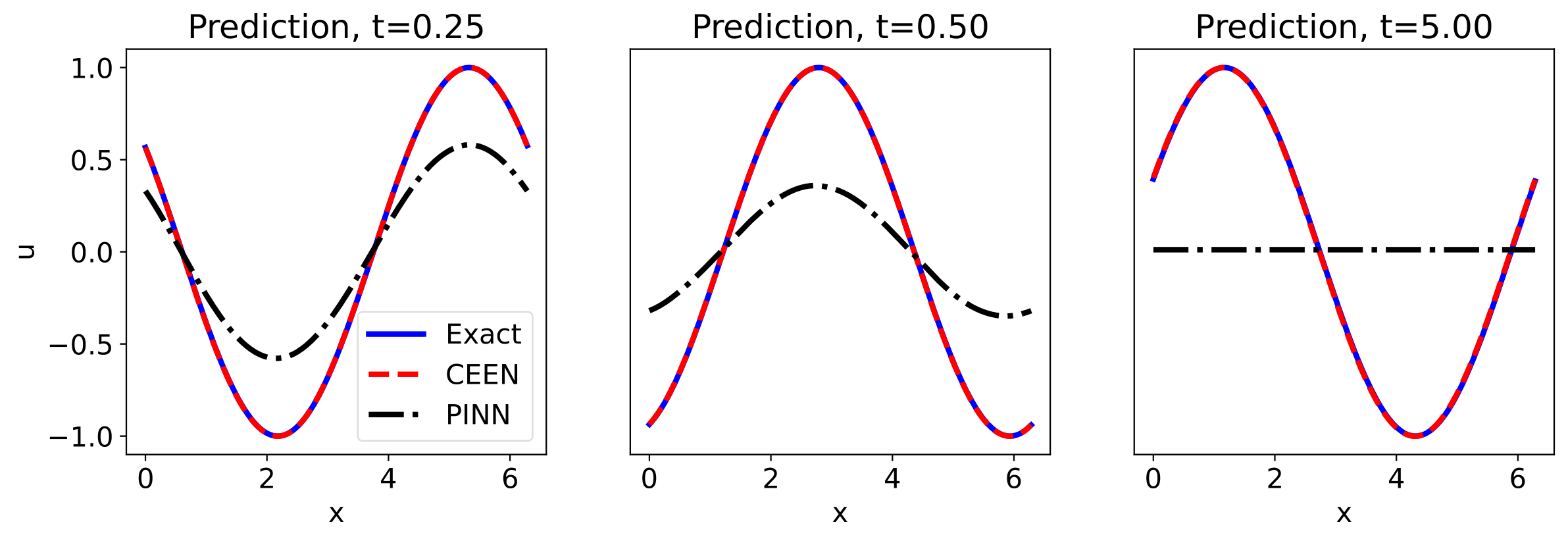}
        \caption{}
    \end{subfigure}
    \caption{\textit{Convection equation:} 
    Comparison of the prediction of \AL~and PINN against the reference
    solution. (a) displays the prediction across the entire spatio-temporal
    domain. (b) displays it at three different times: $t=0.25,0.5,5$. \AL~agrees well
    with the reference in the entire domain while PINN fails to approximate the
    solution.}
    \label{fig:convection_solution}
\end{figure}

In this section, we consider the convection equation of the form:
\begin{equation}
\label{eqn:convection}
\begin{aligned}
u_t + 15 u_x &= 0,\;\;\text{for $t \in [0,5], x \in[0,2\pi],$} \\
u(0,x) &= \sin x, \\
u(t,0) &= u(t,2\pi) \\
u_x(t,0) &= u_x(t,2\pi).
\end{aligned}
\end{equation}

The exact solution for the convection problem can be easily obtained:
\begin{equation} \label{eqn:convection_sol}
    u(x,t) = \sin (x - 15 t).
\end{equation}
We compare two methods, PINN and CEEN, with the exact solution \eqref{eqn:convection_sol}.
In CEEN, we discretize the time domain with the time step size
$h=\text{2.5e-3}$. 
We use $N_x=32$ equidistant points on the spatial domain $\Omega$ as the
collocation points and the tolerance is $\epsilon=\text{1e-5}$.

The solution at each time step is represented using a fully-connected neural network. 
The architecture of the network includes two hidden layers, each containing 15 neurons. 
The chosen values for $K$ and $M$ used are 1 and 10, respectively.
The setting for PINN is established in a similar manner to that described in the previous example.

\begin{figure}
    \centering
    %\begin{subfigure}{0.49\linewidth}
        %\centering 
        \includegraphics[height=4.5cm]{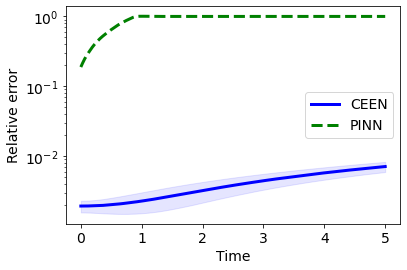}
        %\caption{}
    %\end{subfigure}
    \caption{Relative errors of \AL~ and PINN for the convection equation
    \eqref{eqn:convection}. Shaded area is a standard deviation band for the
    prediction of CEEN obtained by running ten independent simulations.}
    \label{fig:convection_relative_error}
\end{figure}

The results are summarized in Figure \ref{fig:convection_solution}, which shows
good agreement between our prediction and the reference solution.
However, the PINN model falls short, failing to accurately predict the solution
even at an earlier time, such as $t=0.25$, as shown in Figure
\ref{fig:convection_solution}-(b).
Figure \ref{fig:convection_relative_error} illustrates that the relative errors of PINN and FDM are much higher than that of CEEN. 

\begin{figure}
    \begin{subfigure}{0.49\linewidth}
        \centering \includegraphics[height=4.5cm]{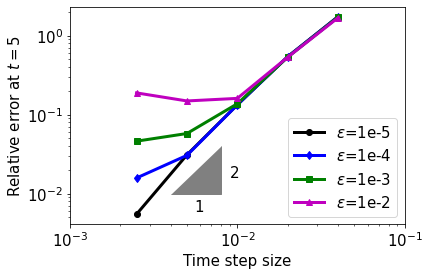}
        \caption{}
    \end{subfigure}
    \begin{subfigure}{0.49\linewidth}
        \centering \includegraphics[height=4.5cm]{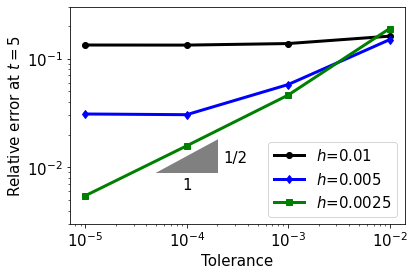}
        \caption{}
    \end{subfigure}
    \caption{Relative error of our method at final time $(T=5)$ for different
    time step sizes $h$ and tolerances $\epsilon$ for the convection equation
    \eqref{eqn:convection}. (a) Time step size vs relative error. 
    The second order of accuracy with respect to the time step size $h$ 
    is achieved for sufficiently small tolerance $\epsilon$. 
    (b) Tolerance vs relative error. 1/2 order of accuracy with respect to the
    tolerance $\epsilon$ is achieved for sufficiently small $h$.}
    \label{fig:convection_error}
\end{figure}

Next, we investigate the relationship between the relative error of the
predicted solution at the final time and factors such as the time step size $h$ or the tolerance
$\epsilon$. 
As derived in \eqref{eqn:error_at_each_time_step}, the error of our method
consists of two terms $e_1(\epsilon,h)=|\Omega|^{1/2}\epsilon^{1/2}h$ and
$e_2(h)=\frac{1}{12}h^3\|\mathcal{N}[u]_{tt}(t_\ast,\cdot)\|_{L^2(\Omega)}$,
representing the training loss and the consistency error, respectively. We
expect that the second order accuracy with respect to $h$ is
attained as $e_1(\epsilon,h)$ has a negligible contribution to the
overall error due to a small value of $\epsilon$. 
Indeed, Figure \ref{fig:convection_error}-(a) confirms this, demonstrating a
second-order accuracy in time for sufficiently small tolerance $\epsilon$. 
Alternatively, we expect that
the error with respect to $\epsilon$ has the order of $1/2$, the
exponent of $\epsilon$ in $e_1(\epsilon,h)$, as $e_2(h)$'s contribution to the
error becomes negligible with a sufficiently small $h$. 
Indeed, Figure \ref{fig:convection_error}-(b) verifies that, for a sufficiently
small time step size $h=0.0025$, our method achieves the error of the order of
$1/2$.

%--------------------------------------------------------------
% SUBSECTION: ALLEN-CAHN
%--------------------------------------------------------------
\subsection{Allen-Cahn equation}

Our final example deals with the Allen-Cahn equation to highlight the capability
of our proposed method in solving phase field models. This equation is commonly
used in the area of reaction-diffusion systems which characterizes the phase
separation process in multi-component alloy systems. The governing equation for
this system can be expressed as follows:
\begin{equation}
\label{eqn:allencahn}
\begin{aligned}
    u_t - 0.0001u_{xx} + 5u^3 - 5u&=0,\;\;\text{for $t \in[0,5], x \in[-1,1]$} \\
u(0,x) &= x^2\cos\pi x, \\
u(t,-1) &= u(t,1), \\
u_x(t,-1) &= u_x(t,1).
\end{aligned}
\end{equation}
To implement our method to solve this equation, we discretize the time domain
with the step size $h=\text{1e-2}$. A fully-connected neural network with two
hidden layers, each containing 32 neurons, is used to represent the solution at
each time step. We use $N_x = 201$ equidistant points sampled points and the
tolerance is $\text{1e-6}$. The chosen values for $K$ and $M$ are
2 and 5, respectively.
The setting for PINN method is established in a similar manner to that
described in the previous examples.

\begin{figure}
    \centering
    \begin{subfigure}{\linewidth}
        \centering \includegraphics[height=4.5cm]{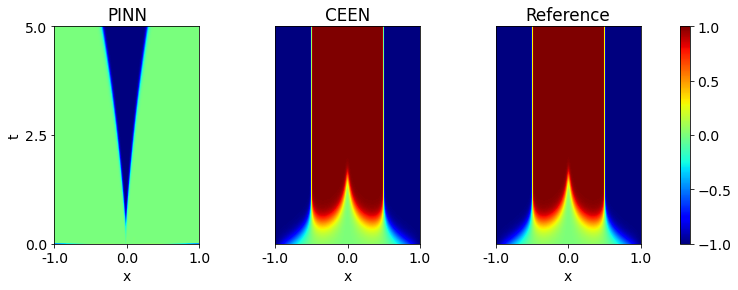}
        \caption{}
    \end{subfigure}
    \centering
    \begin{subfigure}{\linewidth}
        \centering \includegraphics[height=4.5cm]{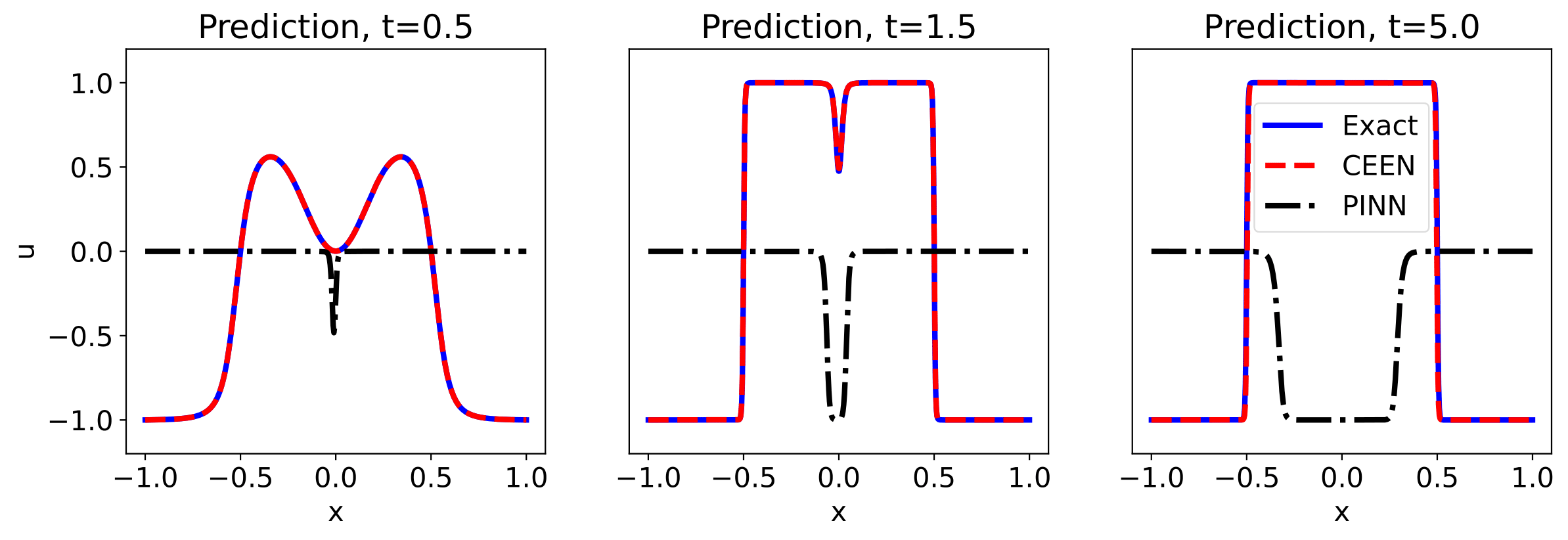}
        \caption{}
    \end{subfigure}
    \caption{\textit{Allen-Cahn equation:} 
    Comparison of the prediction of \AL~and PINN against the reference
    solution. (a) displays the prediction across the entire spatio-temporal
    domain. (b) displays it at three different times: $t=0.5,1.5,5$. \AL~agrees well
    with the reference in the entire domain while PINN fails to approximate the
    solution.}
    \label{fig:allencahn_solution}
\end{figure}

\begin{figure}
    \centering
    %\begin{subfigure}{0.49\linewidth}
        %\centering 
        \includegraphics[height=5cm]{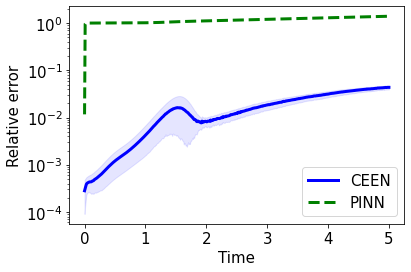}
    %    \caption{}
    %\end{subfigure}
    \caption{Relative errors of \AL~and PINN for the
    Allen-Cahn equation \eqref{eqn:allencahn}. Shaded area is a standard deviation band for the
    prediction of CEEN obtained by running ten independent simulations.}
    \label{fig:allencahn_relative_error}
\end{figure}

Figure \ref{fig:allencahn_solution} shows the comparison of \AL~and PINN against
the reference solution. The prediction of our method agrees very well with the
reference solution over the entire domain while PINN fails to learn the solution
even at earlier time. Our method effectively captures the discontinuity in the
solution. Figure \ref{fig:allencahn_relative_error} show the relative error of
\AL~and PINN over time with \AL~consistently outperforming PINN, thereby underscoring its superior effectiveness.

%--------------------------------------------------------------
%
% SUBSECTION: CONCLUSION
%
%--------------------------------------------------------------
\section{Conclusion}\label{sec:conclusion}

In this study, we have developed and demonstrated a novel sequential learning
strategy, termed as \AL, designed to solve long-time integration of PDEs. We
achieved this by dividing the time domain into non-overlapping subintervals,
assigning a unique neural network to each subinterval, and constructing a loss
function based on the integral form of PDEs within these subintervals. Then, we
enforced a sequential training, beginning with the initial time step. 

Throughout the application of our method to various PDE problems, including the
Korteweg-de Vries (KdV) equation, Schrödinger equation, convection equation, and
Allen-Cahn equation, we have consistently observed superior performance compared
to the PINN. In particular, our method demonstrated remarkable agreement with reference
solutions across entire spatio-temporal domains. Furthermore, the parallel
training showed a significant computational speedup, making it an efficient choice
for complex problems.
We have also demonstrated the second-order accuracy with respect to the time
step and the half-order accuracy with respect to the tolerance. 
These findings validate the theoretical error analysis and
provide insights into the accuracy of our method. However, more comprehensive error
analysis will be left for future work. 

While we have used the trapezoidal rule as one of many potential
linear-multistep methods for formulating the loss function, the flexibility of
our approach allows for the straightforward incorporation of any other
linear-multistep methods, such as Adams-Moulton methods, for loss function
construction. This extensibility opens up avenues for further research.
High-order methods, for example, are anticipated to accommodate larger time
steps and ensure high-order accuracy, which will be investigated in future work.

\section*{Acknowledgments}

This work was supported by the National Research Foundation of Korea
[NRF-2021R1C1C1007875].

%--------------------------------------------------------------
%
% SECTION: APPENDIX A.
%
%--------------------------------------------------------------
\section*{Appendix A. Comparison between CEEN and classical numerical method}\label{sec:appendix_a}

In this Appendix, we conduct a comparison between CEEN and classical numerical
methods, focusing on the convection equation discussed in Section~\ref{sec:convection}. 
The presence of a closed-form solution for this equation enables the precise
calculation of accuracy. We select the Finite Difference Method (FDM) as the
classical numerical method for comparison, using the same
time-integrator, the Trapezoidal discretization, as employed in CEEN to ensure a
fair comparison. For spatial discretization, a centered difference scheme is
utilized.
It is noteworthy that the selected FDM is absolute stable~\cite{durran} and
achieves second-order accuracy both in space and time.

For CEEN, we choose the same training configuration as detailed in
Section~\ref{sec:convection}. The parameters used for this comparison are
summarized in Table~\ref{tab:convection_params}.

\begin{table}[H]
    \centering
    \caption{CEEN parameters for the convection equation}
    \label{tab:convection_params}    
    \begin{tabular}{||c|c||c|c||}
        \hline
         & Architecture & & Training\\
        \hline
        NN shape & [2,15,15,1] & optimizer & Adam\\
        activation & $\sin$ & learning rate & $10^{-3}$\\
        $M$ & 10 & optimizer parameters & $\beta = (0.999,0.999)$\\
        $K$ & 1 & $\epsilon$ & $10^{-5}$\\
        & & Maximum iteration& $100,000$\\
        \hline
    \end{tabular}
\end{table}

\begin{figure}[H]
    \centering
    \begin{subfigure}{0.33\linewidth}
        \centering \includegraphics[height=3.5cm]{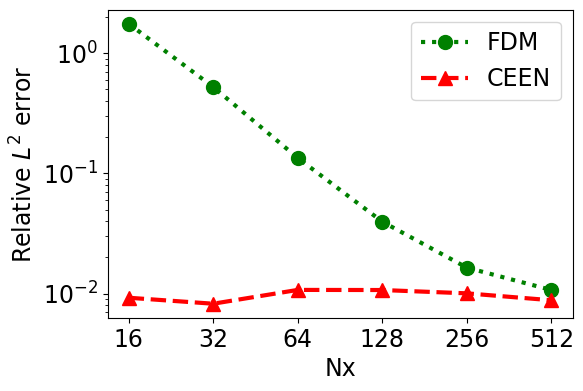}
        \caption{}
    \end{subfigure}
    \centering
    \begin{subfigure}{0.66\linewidth}
        \centering \includegraphics[height=3.5cm]{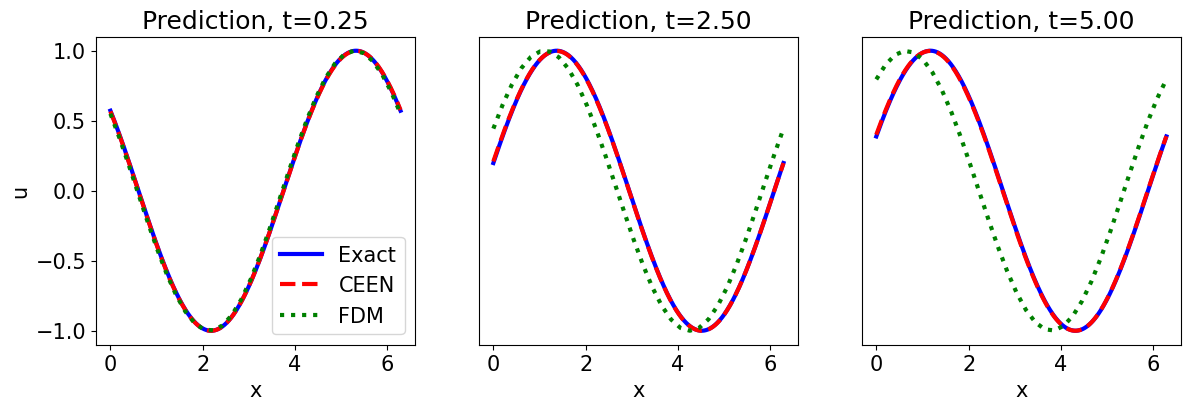}
        \caption{}
    \end{subfigure}
    \caption{
    Comparison of CEEN and finite difference method (FDM).
    (a) Relative errors at the final time $t=5$ vs. $N_x$.
    (b) Predictions of solution with $N_x = 32$ at three different times.
    }
    \label{fig:convection_ceen_fdm1}
\end{figure}

Firstly, we fix the number of time steps as $N_t=2001$ (equivalent to $dt=0.0025$) and vary the number
of spatial grids ($N_x$) to analyze how the error behaves with respect to $N_x$. 
Figure~\ref{fig:convection_ceen_fdm1}-(a) illustrates the relative
$L_2$ error at the final time $t=5$ for both FDM and CEEN. 
FDM exhibits second-order accuracy with respect to $N_x$ due to its spatial
discretization, achieving an error of $10^{-2}$ with $N_x=512$. 
Conversely, the error of CEEN shows independence of the number of spatial grids,  
thanks to its precise computation of spatial derivatives through automatic
differentiation, as discussed in Section~\ref{sec:analysis}. 
Notably, CEEN achieves a lower error with just $N_x=16$ compared to 
FDM even with $N_x=512$. Figure~\ref{fig:convection_ceen_fdm1}-(b)
compares the prediction of CEEN and FDM with $N_x=32$ at three different times:
 $t=0.25, 2.5$ and $5$. FDM successfully approximates the solution at the early
time $=0.25$, but struggles to accurately approximate the solution at later times. However,
CEEN predicts the solution accurately at all times.

\begin{figure}[H]
    \centering
    \begin{subfigure}{0.49\linewidth}
        \centering \includegraphics[height=5cm]{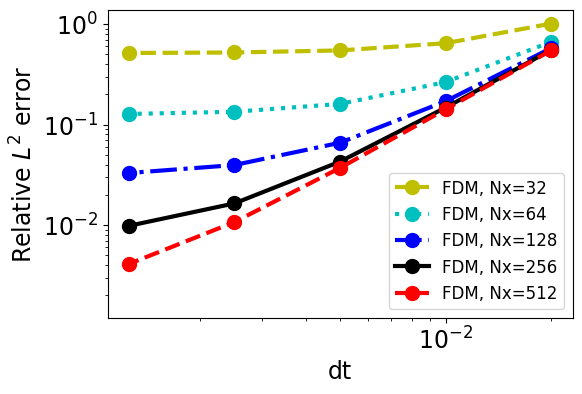}
        \caption{}
    \end{subfigure}
    \centering
    \begin{subfigure}{0.49\linewidth}
        \centering \includegraphics[height=5cm]{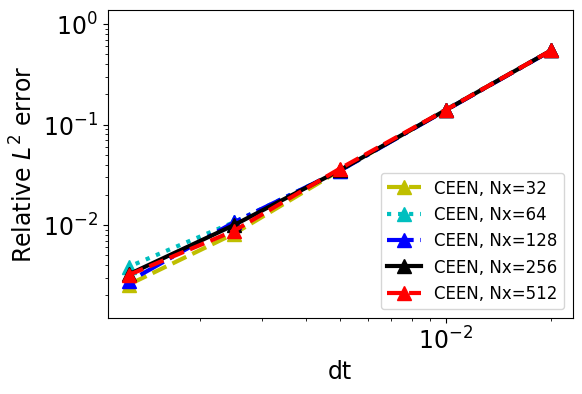}
        \caption{}
    \end{subfigure}
    \caption{
    Relative $L_2$ errors vs. $dt$ at the final time $t=5$ for FDM (a) and
    CEEN (b). 
    }
    \label{fig:comparison_fdm_Nt}
\end{figure}

Secondly, we present a quantitative analysis comparing the relative errors of FDM and
CEEN with respect to the time step size $dt$ (equivalent to the number of temporal
steps $N_t$). Figure~\ref{fig:comparison_fdm_Nt} illustrates the relative $L_2$
error at the final time $t=5$ for both FDM (left panel) and CEEN (right panel). 
For FDM, when $N_x$ is low, we observe that the error decreases and then
plateaus as $dt$ decreases, indicating that the error due to spatial
discretization becomes predominant. However, with large number of spatial grids
such as $N_x=512$, the error
exhibits second-order accuracy in relation to $dt$, a direct result of
the Trapezoidal rule. This behavior illustrates a well-recognized phenomenon in
traditional numerical methods, where the error is influenced by the ratio of
$dt$ and $dx$.
Conversely, the error for CEEN consistently achieves second-order
accuracy in relation to $dt$, independent of the number of spatial grids, as
discussed in Section~\ref{sec:analysis}.

%--------------------------------------------------------------
%
% SECTION: APPENDIX B.
%
%--------------------------------------------------------------
\section*{Appendix B. Computational cost and accuracy}\label{sec:appendix_b}

Here we investigate the computational cost of CEEN compared with that of PINN. 
Additionally, we explore the effects of parallelization on the computational
efficiency and accuracy within the CEEN framework. For consistency and accurate
comparison, we employ identical hyperparameter configurations for both CEEN and
PINN, as detailed in Section~\ref{sec:numerical}, across all four problems 
under consideration. The specific hyperparameter settings for CEEN are
outlined in Table~\ref{tab:appb}.

\begin{table}[H]
    \centering
    \caption{CEEN hyperparameters for four PDE problems.}
    \label{tab:time}    
    \begin{tabular}{l|r|r|r|r|r|r}
        \hline
        \multicolumn{1}{c|}{\multirow{2}{*}{Example}} &
          \multicolumn{6}{c}{Settings} \\
          \cline{2-7}
        \multicolumn{1}{c|}{} &
          \multicolumn{1}{c|}{Architecture} &
          \multicolumn{1}{c|}{Activation} &
          \multicolumn{1}{c|}{$N_x$} &
          \multicolumn{1}{c|}{$N_t$} &
          \multicolumn{1}{c|}{$\epsilon$} &
          \multicolumn{1}{c}{$K$} \\
          \hline
        KdV & [5,16,16,1] & tanh &151 & 501 & $10^{-4}$ & 3 \\
        Schr{\"o}dinger & [3,16,16,2] & tanh &151 & 101 & $10^{-5}$ & 2 \\
        Convection & [2,15,15,1] & tanh & 101 & 1001 & $10^{-5}$ & 1 \\
        Allen-Cahn & [1,32,32,1] & tanh & 201 & 501 & $10^{-5}$ & 2 \\
        \hline
    \end{tabular}
    \label{tab:appb}
\end{table}

%--------------------------------------------------------------
% SUB_SECTION: APPENDIX B.1
%--------------------------------------------------------------
\subsection*{B.1. Computational cost of PINN and CEEN}\label{sec:compcost}

\begin{figure}[H]
    \centering
    \begin{subfigure}{0.9\linewidth}
        \centering \includegraphics[height=6.0cm]{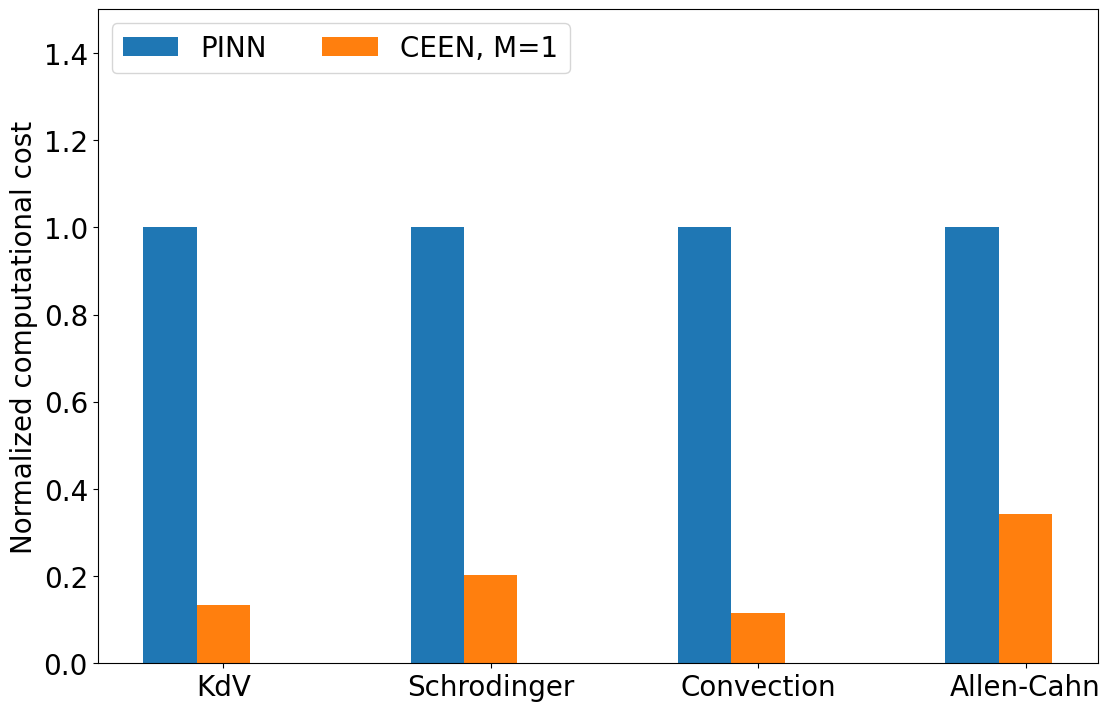}
    \end{subfigure}
    \caption{Comparison of computational cost between PINN and
    CEEN ($M=1$). y-axis: the ratio of the computation time of CEEN to that of
    PINN. For all four problems, the computational cost of PINN is much larger
    than that of CEEN. }
    \label{fig:trainingtime_PINN_CEEN}
\end{figure}

In this Appendix, we examine the computational efficiency of the
CEEN in a non-parallelized setup ($M=1$) compared to PINN across four case studies
presented in Section~\ref{sec:numerical}. CEEN involves training
multiple neural networks (NNs) sequentially, with each NN designed to model the
solution at a given time step. This modular design allows for each NN in the
CEEN framework to be more compact, as it needs to accurately capture solution
dynamics over a small time interval.
On the other hand, PINN employs a single but significantly larger NN to approximate the solution
over the entire spatio-temporal domain. Training a smaller NN, as in the CEEN
approach, is inherently faster than training a larger one. 
Even though CEEN incorporates many NNs,
dictated by the number of time steps and the total duration of simulation,
Figure~\ref{fig:trainingtime_PINN_CEEN} demonstrates that CEEN surpasses PINN in
computational efficiency for all considered problems. Consequently, CEEN not only
reduces computational costs in comparison to PINN but also achieves 
higher accuracy in long-term simulations of diverse PDE problems, where
traditional PINN method fails.

%--------------------------------------------------------------
% SUB_SECTION: APPENDIX B.2
%--------------------------------------------------------------
\subsection*{B.2. Effects of parallelization on the computational efficiency
and accuracy in CEEN}

\begin{figure}[H]
    \centering
    \begin{subfigure}{0.9\linewidth}
        \centering \includegraphics[height=6.0cm]{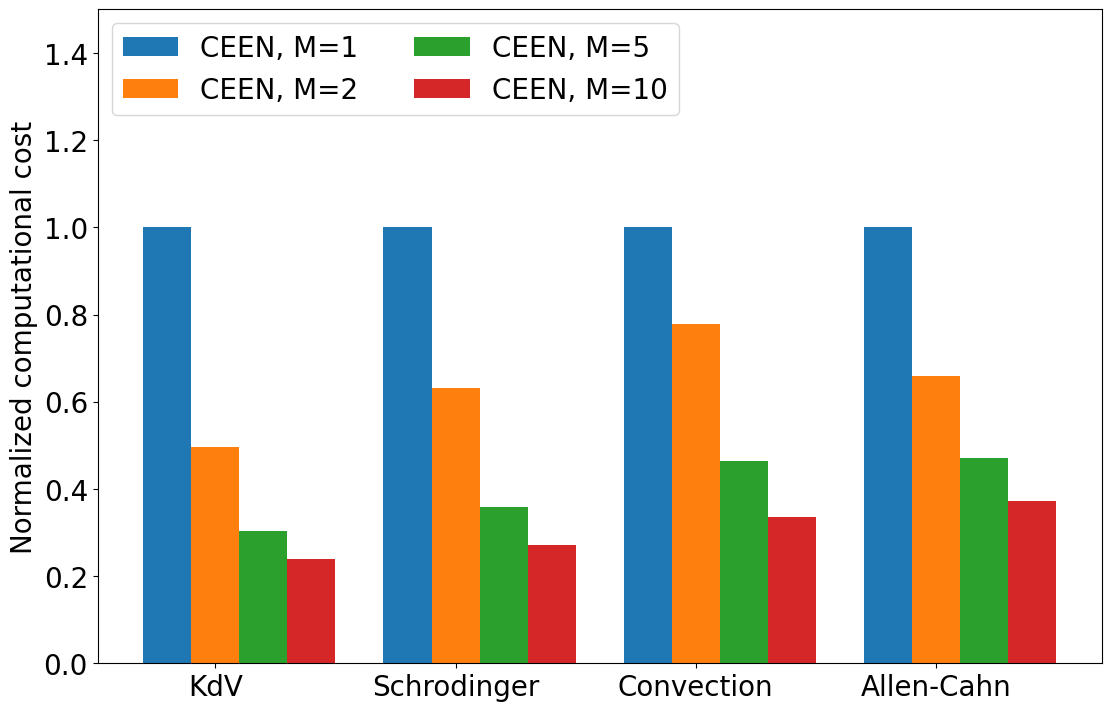}
    \end{subfigure}
    \caption{Comparison of computational cost across different
    $M$ values in CEEN. y-axis: the ratio of the computation time for the cases
    in parallel ($M>1$) to that for the case without parallelization ($M=1$).}
    \label{fig:ceen_trainingtime_bar}
\end{figure}

We investigate the computational efficiency of the CEEN algorithm with
parallelization presented in Algorithm~\ref{alg:tren}. We examine the
computation times needed
to solve each PDE problem in Section~\ref{sec:numerical}, utilizing
parallelization levels of $M=2, 5,$ and $10$,
 and compare these results to the computation time with $M=1$. 
Figure~\ref{fig:ceen_trainingtime_bar} demonstrates that parallelization reduces computation time. 
Moreover, it is crucial to highlight that the accuracy of the parallel CEEN
, documented in Table~\ref{tab:time_rel_error}, remains consistently comparable and
does not deteriorate with increasing values of $M$. This consistency in accuracy
underscores the efficacy of the parallel CEEN algorithm, affirming its
robustness and effectiveness in computational performance.

\begin{table}[H]
    \centering
    \caption{The relative $L^2$ error at final time for four PDE problems across
    different values of $M$}
    \begin{tabular}{l|r|r|r|r}
        \hline
        \multicolumn{1}{c|}{\multirow{2}{*}{Example}} &
          \multicolumn{4}{c}{M} \\
          \cline{2-5}
        \multicolumn{1}{c|}{} &
          \multicolumn{1}{c|}{1} &
          \multicolumn{1}{c|}{2} &
          \multicolumn{1}{c|}{5} &
          \multicolumn{1}{c}{10} \\
          \hline
        KdV & 2.159e-2 & 2.433e-2 & 1.700e-2 & 2.600e-2 \\
        Schr{\"o}dinger  & 3.584e-3 & 3.573e-3 & 5.080e-3 & 4.129e-3 \\
        Convection & 9.232e-3 & 9.293e-3 & 9.199e-3 & 9.204e-3 \\
        Allen-Cahn  & 4.200e-2 & 3.325e-2 & 2.229e-2 & 2.049e-2 \\
        \hline
    \end{tabular}
    \label{tab:time_rel_error}
\end{table}

\bibliography{references}

\end{document}